%% file: paper.tex
\def\year{2021}\relax
\documentclass[letterpaper]{article} 
\usepackage{aaai21}  
\usepackage{times}  
\usepackage{helvet} 
\usepackage{courier}  
\usepackage[hyphens]{url}  
\usepackage{graphicx} 
\usepackage{natbib}  
\usepackage{caption} 
\title{Adversarial Training and Provable Robustness: A Tale of Two Objectives}
\author {
    Jiameng Fan , Wenchao Li\\
}

\affiliations{
    Department of Electrical and Computer Engineering \\
    Boston University, Boston \\
    \{jmfan, wenchao\}@bu.edu
}

\input{math_commands.tex}

\usepackage{booktabs}       
\usepackage{amsfonts}       
\usepackage{nicefrac}       
\usepackage{microtype}      
\usepackage{graphicx, subfig, multirow, adjustbox}
\usepackage{amsthm}
\usepackage{algorithm}
\usepackage{algpseudocode}
\usepackage[table,xcdraw]{xcolor}
\usepackage{threeparttable}
\usepackage{enumitem}
\usepackage{xspace}
\usepackage{cuted}
\usepackage{multicol}

\newtheorem{theorem}{Theorem}
\newtheorem{lemma}{Lemma}
\newtheorem{assumption}{Assumption}

\newcommand{\toolname}{\textsl{AdvIBP}\xspace}
\newcommand{\crowntoolname}{\textsl{AdvCROWN-IBP}\xspace}

\usepackage{xcolor}

\begin{document}

\maketitle

\begin{abstract}
  We propose a principled framework that combines adversarial training and provable robustness verification for training certifiably robust neural networks.
  We formulate the training problem as a joint optimization problem with both empirical and provable robustness objectives and develop a novel gradient-descent technique that can eliminate bias in stochastic multi-gradients. We perform both theoretical analysis on the convergence of the proposed technique and experimental comparison with state-of-the-arts. Results on MNIST and CIFAR-10 show that our method can consistently match or outperform prior approaches for provable $\evl_\infty$ robustness. Notably, we achieve 6.60\% verified test error on MNIST at $\epsilon=0.3$, and 66.57\% on CIFAR-10 with $\epsilon=8/255$.
\end{abstract}

\section{Introduction}
\input{tex/introduction.tex}
\label{sec:introduction}


\section{Background}
\input{tex/background.tex}
\label{sec:background}

\section{Methodology}
\input{tex/methodology.tex}
\label{sec:methodology}

\section{Experiment}
\input{tex/experiment.tex}
\label{sec:experiment}

\section{Conclusion}
\input{tex/conclusion.tex}
\label{sec:conclusion}

\section*{Acknowledgement}
We gratefully acknowledge the support from NSF grant 1646497 and ONR grant N00014-19-1-2496.
We also thank K. Wardega, W. Zhou, P. Kiourti and F. Fu at Boston University for their helpful discussion.

\section*{Ethics Statement}


Broad acceptance and adoption of large-scale deployments of deep learning systems rely critically on their trustworthiness which, in turn, depends on the ability to assess and demonstrate the safety of such systems.
Concerns like adversarial robustness already arise with today's deep learning systems and those that may be exacerbated in the future with more complex systems.
Our research has the potential to enable the efficient training of robust deep learning systems. It can help unlock deep learning applications that are currently not deployable due to safety, robustness or resource concerns.
These applications range from autonomous driving to mobile devices, and can benefit the society at large.

\bibliography{jiameng}

\clearpage

\appendix
\begin{strip}
\section{Appendix}
\end{strip}
\input{tex/appendix.tex}
\label{sec:appendix}

\end{document}

%% file: math_commands.tex
\usepackage{amsmath,amsfonts,bm}

\def\Figref#1{Figure~\ref{#1}}

\def\eqref#1{equation~\ref{#1}}

\def\Algref#1{Algorithm~\ref{#1}}

\def\1{\bm{1}}

\def\vb{{\bm{b}}}
\def\vc{{\bm{c}}}

\def\vh{{\bm{h}}}

\def\vm{{\bm{m}}}

\def\vv{{\bm{v}}}

\def\vx{{\bm{x}}}

\def\vz{{\bm{z}}}

\def\evl{{l}}

\def\mW{{\bm{W}}}

\DeclareMathAlphabet{\mathsfit}{\encodingdefault}{\sfdefault}{m}{sl}
\SetMathAlphabet{\mathsfit}{bold}{\encodingdefault}{\sfdefault}{bx}{n}

\def\gB{{\mathcal{B}}}

\def\gL{{\mathcal{L}}}

\def\gX{{\mathcal{X}}}
\def\gY{{\mathcal{Y}}}

\def\sS{{\mathbb{S}}}

\newcommand{\E}{\mathbb{E}}

\DeclareMathOperator*{\argmax}{arg\,max}

%% file: tex/introduction.tex
Vulnerability of deep neural networks to adversarial examples~\citep{szegedy2014intriguing, goodfellow2015explaining} has spurred the development of training methods for learning more robust models~\citep{wong2018provable,gowal2018effectiveness,zhang2019towards,balunovic2020adversarial}.
\citet{madry2017towards} show that adversarial training can be formulated as a minimax robust optimization problem as in (\ref{eq: minimax}). Given a model $f_\theta$, loss function $\gL$, and training data distribution $\gX$, the training algorithm aims to minimize the loss whereas the adversary aims to maximize the loss within a neighborhood $\sS(\vx, \epsilon)$ of each input data $\vx$ as follows:
\begin{equation}
    \min_{\theta} E_{(\vx, y)\in \gX} \left[\max_{\vx' \in \sS(\vx, \epsilon)} \gL(f_\theta(\vx'), y) \right]
    \label{eq: minimax}
\end{equation}
In general, the inner maximization is intractable. Most existing techniques focus on finding an approximate solution. 
There are two main approaches to approximate the inner loss (henceforth referred to as \textit{robust loss}). One direction is to generate adversarial examples to compute a lower bound of robust loss. The other is to compute an upper bound of robust loss by over-approximating the model outputs. We distinguish these two families of techniques below.

\textbf{Adversarial training.}
To improve adversarial robustness, a natural idea is to augment the training set with adversarial examples~\citep{kurakin2017adversarial}. Using adversarial examples to compute the training loss yields a lower bound of \textit{robust loss}, henceforth referred to as \textit{adversarial loss}. \citet{madry2017towards} propose to use projected gradient descent (PGD) to compute the adversarial loss and train the neural network by minimizing this loss. 
Networks trained using this method can achieve state-of-art test accuracy under strong adversaries~\citep{carlini2017towards, wang2018mixtrain}. 
More recently, \citet{wong2020fast} showed that fast gradient sign method (FGSM)~\citep{goodfellow2015explaining} with random initialization can be used to learn robust models faster than PGD-based adversarial training. 
In term of efficiency, FGSM-based adversarial training is comparable to regular training.
While adversarial training can produce networks robust against strong attacks, minimizing the adversarial loss alone cannot guarantee that (\ref{eq: minimax}) is minimized.
In addition, it cannot provide rigorous guarantees on the robustness of the trained networks. 

\textbf{Provable robustness.}
Verification techniques~\citep{katz2017reluplex, dvijotham2018dual,ruan2018reachability,raghunathan2018semidefinite,prabhakar2019abstraction}, on the other hand, can be used to compute a certified upper bound of \textit{robust loss} (henceforth referred to as \textit{abstract loss}).
Given a neural network, a simple way to obtain this upper bound is to propagate value bounds across the network, also known as interval bound propagation (IBP)~\citep{mirman2018differentiable, gowal2018effectiveness}.
Techniques such as CROWN~\citep{zhang2018efficient}, DeepZ~\citep{singh2018fast}, MIP~\citep{tjeng2019evaluating} and RefineZono~\citep{singh2019boosting}, can compute more precise bounds, but also incur much higher computational costs.
Building upon these upper bound verification techniques, approaches such as DIFFAI~\citep{mirman2018differentiable} construct a differentiable \textit{abstract loss} corresponding to the upper bound estimation and incorporate this loss function during training. 
However, \citet{gowal2018effectiveness} and \citet{zhang2019towards} observe that a tighter approximation of the upper bound does not necessarily lead to a network with low robust loss. They show that IBP-based methods can produce networks with state-of-the-art certified robustness. 
More recently, 
COLT~\citep{balunovic2020adversarial}
proposed to combine adversarial training and zonotope propagation.
Zonotopes are a collection of affine forms of the input variables and intermediate vector outputs in the neural network.
The idea is to train the network with the so-called latent adversarial examples which are adversarial examples that lie inside these zonotopes.

\begin{table*}[htpb]
\caption{Comparison of different methods for training robust neural networks. We highlight the loss function used in each method. If there is an \textit{abstract loss} used in training or post-training verification, we also list the corresponding verification method. We categorize the methods along five dimensions, with $\checkmark$ indicating a desirable property or an explicit consideration.}
\begin{adjustbox}{width=0.95\textwidth, center}
\begin{threeparttable}
\begin{tabular}{@{}cccccccc@{}}
\toprule
Method & Loss & \begin{tabular}[c]{@{}c@{}}$\textit{Abstract loss}$\end{tabular} & Efficiency\tnote{1}
& \begin{tabular}[c]{@{}c@{}}Empirical\\ Robustness\end{tabular} & \begin{tabular}[c]{@{}c@{}}Provable\\ Robustness\end{tabular} & \begin{tabular}[c]{@{}c@{}}No weight\tnote{2}\\ tuning/scheduling\end{tabular} \\ \midrule
Baseline & $\textit{regular loss}$ & \multicolumn{1}{c|}{n/a} & \multicolumn{1}{c|}{$\checkmark$}
& \multicolumn{1}{c|}{} & \multicolumn{1}{c|}{} & n/a \\
FGSM~\shortcite{goodfellow2015explaining} & $\textit{adversarial loss}$ & \multicolumn{1}{c|}{n/a} & \multicolumn{1}{c|}{$\checkmark$}
& \multicolumn{1}{c|}{$\checkmark$} & \multicolumn{1}{c|}{} & n/a \\
FGSM+random init~\shortcite{wong2020fast} & $\textit{adversarial loss}$ & \multicolumn{1}{c|}{n/a} & \multicolumn{1}{c|}{$\checkmark$} 
& \multicolumn{1}{c|}{$\checkmark$} & \multicolumn{1}{c|}{} & n/a \\
PGD~\shortcite{madry2017towards} & $\textit{adversarial loss}$ & \multicolumn{1}{c|}{n/a} & \multicolumn{1}{c|}{}
& \multicolumn{1}{c|}{$\checkmark$} & \multicolumn{1}{c|}{} & n/a \\
COLT~\shortcite{balunovic2020adversarial} & $\textit{latent adversarial loss}$ & \multicolumn{1}{c|}{RefineZono\tnote{3}} & \multicolumn{1}{c|}{} 
& \multicolumn{1}{c|}{$\checkmark$} & \multicolumn{1}{c|}{$\checkmark$} & n/a \\
DIFFAI~\shortcite{mirman2018differentiable} & $\textit{abstract loss}$\tnote{4} & \multicolumn{1}{c|}{DeepZ} & \multicolumn{1}{c|}{$\checkmark\kern-1.1ex\raisebox{.7ex}{\rotatebox[origin=c]{125}{--}}$\tnote{5}}
& \multicolumn{1}{c|}{} & \multicolumn{1}{c|}{$\checkmark$} & n/a  \\
CROWN-IBP~\shortcite{zhang2019towards} & $\textit{regular loss} {+} \textit{abstract loss}$ & \multicolumn{1}{c|}{CROWN + IBP} & \multicolumn{1}{c|}{$\checkmark$}
& \multicolumn{1}{c|}{} & \multicolumn{1}{c|}{$\checkmark$} &  \\
IBP method~\shortcite{gowal2018effectiveness} & $\textit{regular loss} {+} \textit{abstract loss}$ & \multicolumn{1}{c|}{IBP} & \multicolumn{1}{c|}{$\checkmark$}
& \multicolumn{1}{c|}{} & \multicolumn{1}{c|}{$\checkmark$} &  \\
\textbf{\toolname} & \textbf{$\textit{adversarial loss} {+} \textit{abstract loss}$} & \multicolumn{1}{c|}{\textbf{IBP}} & \multicolumn{1}{c|}{$\checkmark$}
& \multicolumn{1}{c|}{$\checkmark$} & \multicolumn{1}{c|}{$\checkmark$} & $\checkmark$ \\ \bottomrule
\end{tabular}
\begin{tablenotes}
    \item[1] The efficiency baseline is the training time for each epoch during regular training. $\checkmark$ represents the training time is comparable to the baseline.
    \item[2] The weights here represent the weights for the different losses if there are multiple of them.
    \item[3] RefineZono is not used to construct an abstract loss. Instead, it is used to generate latent adversarial examples and for post-training verification.
    \item[4] In their experiments, DIFFAI shows that adding regular loss with a fixed weight can achieve better performance.
    \item[5] DIFFAI can also use IBP for training and verification for improved efficiency. However, the best robustness results are achieved using DeepZ.
\end{tablenotes}
\end{threeparttable}
\end{adjustbox}
\label{tab:summary}
\end{table*}

\textbf{This work: a principled framework for combining adversarial loss and abstract loss.}
We first start with the observation that there is a substantial gap between the provable robustness obtained from state-of-art verification tools and the empirical robustness of the same network against strong adversary in large-scale models.
In this paper, we propose to bridge this gap
by marrying the strengths of adversarial training and provable bound estimation techniques.
Minimizing \textit{adversarial loss} and minimizing \textit{abstract loss} can be viewed as bounding the true \textit{robust loss} from two ends. 
We argue that simultaneously reducing both losses is more likely to produce a network with good empirical and provable robustness.
From an optimization perspective, this amounts to an optimization problem with two objectives and can be solved using gradient descent methods if both objectives are semi-smooth.
The challenge is how to balance the minimization of these two objectives during training. 
In particular, computing the gradient based on a weighted-sum of the objectives can result in biased gradients.
Inspired by the work on moment estimates~\citep{kingma2014adam}, we propose a novel joint training scheme to compute the weights adaptively and minimize the joint objective with unbiased gradient estimates.
For efficient training, we instantiate our framework in a tool called \toolname, which uses FGSM and random initialization for computing the adversarial loss and IBP for computing the abstract loss. 
We validate our approach on a set of commonly used benchmarks demonstrate and demonstrate that \toolname can learn provably robust neural networks that match or outperform state-of-art techniques.
We summarize and compare the key features of prior methods and \toolname in Table~\ref{tab:summary}.

\noindent
\textbf{Main contributions.} In short, our key contributions are:
\begin{itemize}
	\item A novel framework for training provably robust deep neural networks. The framework marries the strengths of adversarial training and provable upper bound estimation in a principled way.
    \item A novel gradient descent method for two-objective optimization that uses moment estimates to address the issue of bias in stochastic multi-gradients. 
    We also perform theoretical analysis of the proposed method.
    \item Experiments on the MNIST and CIFAR-10 datasets show the proposed method can achieve state-of-the-art performance for networks with provable robustness guarantees.
\end{itemize}

%% file: tex/background.tex
In this paper, we consider an adversary who can perturb an input $\vx {\in} \gX$ from a data distribution $\gX$ arbitrarily within a small $\epsilon$ neighborhood of the input.
In the case of $\evl_\infty$ perturbation, which we experiment with later, we define the allowable adversarial input set as $\sS(\vx, \epsilon){=}\{\vx' | \Vert \vx' {-} \vx \Vert_\infty {\le} \epsilon \}$.

We define a $L$-layer neural network parameterized by $\theta$ as a function $f_\theta$ recursively as:
\begin{align*}
	f_\theta(\vx) {=} \vz^{(L)}, \ \ \vz^{(l)} {=} \mW^{(l)}\vh^{(l-1)} {+} \vb^{(l)},\ \
	\vh^{(l)} {=} \sigma^{(l)}(\vz^{(l)})
\end{align*}
where
$l \in \{1, {\cdots}, L{-}1\}$,
$\vz$ represent the pre-activation neuron values, $\vh$ represent post-activation neuron values and $\sigma$ is an element-wise activation function. We denote $h_{\theta_l}^{(l)}$ the mapping applied at layer $l$ with parameter $\theta_l$ and the network can be represented as $f_\theta = h_{\theta_L}^{(L)} \circ h_{\theta_{L-1}}^{(L-1)} \cdots \circ h_{\theta_1}^{(1)}$.

In classification, the provable robustness seeks for the lower bounds of the margins between the ground-truth logit and all other classes.
Let vector $\vm$ be the margins between the ground-truth class and all other classes.
Each element in $\vm$ is a linear combination of the output~\citep{wong2018provable}:
$
\vc^T f_\theta(\vx)
$, where $\vc$ is set to compute the margin.
We define the lower bound of $\vm$ in $\sS(\vx, \epsilon)$ as $\underline{\vm}(\vx, \epsilon;\theta)$. When all elements of $\underline{\vm}(\vx, \epsilon;\theta) {>} 0$, $\vx$ is verifiably robust for any perturbation with $\evl_\infty$-norm less than $\epsilon$.

\textbf{Interval bound propagation (IBP).}
Interval bound propagation uses a simple bound propagation rule. For the input layer we define element-wise upper and lower bound for $\vx$, $z^{(l)}$ and $\vh^{(l)}$ as $\vx_L {\le} \vx {\le} \vx_U$, $\underline{\vz}^{(l)} {\le} \vz^{(l)} {\le} \overline{\vz}^{(l)}$  and $\underline{\vh}^{(l)} {\le} \vh^{(l)} {\le} \overline{\vh}^{(l)}$. For affine layers, we have:
\begin{equation*}
\begin{array}{cc}
    \overline{\vz}^{(l)} {=} \mW^{(l){-}} {\cdot} \underline{\vh}^{(l{-}1)} {+} \mW^{(l){+}} {\cdot} \overline{\vh}^{(l{-}1)} {+} \vb^{(l)}, \\
    \underline{\vz}^{(l)} {=} \mW^{(l){-}} {\cdot} \overline{\vh}^{(l{-}1)} {+} \mW^{(l){+}} {\cdot} \underline{\vh}^{(l{-}1)} {+} \vb^{(l)}
\end{array}
\end{equation*}
where $\mW^{(l){-}} {=} \min(0, \mW^{(l)})$ and $\mW^{(l){+}} {=} \max(0, \mW^{(l)})$. Note that $\overline{\vh}^{(0)} {=} \vx_U$ and $\underline{\vh}^{(0)} {=} \vx_L$. For monotonic increasing activation functions $\sigma$, we have $\overline{\vh}^{(l)} = \sigma(\overline{\vz}^{(l)})$ and $\underline{\vh}^{(l)} = \sigma(\underline{\vz}^{(l)})$.

We define $\underline{\vm}_{\text{IBP}}(\vx, \epsilon; \theta)$ as the lower bound of the margin obtained by IBP which is an underapproximation of $\underline{\vm}(\vx, \epsilon; \theta)$. More generally, we use $\underline{\vm}_\textit{abstract}(\vx, \epsilon; \theta)$ as the lower bound of the margin obtained by abstract methods. When $\underline{\vm}_{\textit{abstract}}(\vx, \epsilon; \theta) {\ge} 0$, $\vx$ is verifiably robust by the abstract method for any perturbation with $\evl_\infty$-norm less than $\epsilon$. Additionally, \citet{wong2018provable} showed that for cross-entropy (CE) loss:
\begin{equation}
    \max_{\vx' \in \sS(\vx, \epsilon)} \gL(f_\theta(\vx'), y) \le \gL(-\underline{\vm}_{\textit{abstract}}(\vx, \epsilon; \theta), y; \theta)
    \label{eq: adv_upper}
\end{equation}
IBP or other abstract methods gives a tractable upper bound of the inner-max in (\ref{eq: minimax}) and we refer it as \textit{abstract loss}. 
In practice, solely minimizing \textit{abstract loss} can be unstable and hard to tune~\citep{mirman2018differentiable, gowal2018effectiveness}.
To mitigate this instability, prior works~\citep{mirman2018differentiable, gowal2018effectiveness,zhang2019towards} propose to stabilize the minimization of the abstract loss by adding normal regular loss in the objective. More specifically, the new objective can be formed as follows:
\begin{equation}
    \gL(\theta) = \kappa_1 \gL(f_\theta(\vx),y) + \kappa_2 \gL(-\underline{\vm}_\textit{abstract};y;\theta)
    \label{eq:prior_obj}
\end{equation}
The coefficients $\kappa_1$ and $\kappa_2$ are hand-tuned to balance the minimization between regular loss and abstract loss. The goal is to improve the robustness of the trained model while avoiding the instability caused by loose abstract loss with respect to the true robust loss.
Among different abstract methods, computing IBP bounds only requires two simple forward passes through the network and is thus computationally efficient. 
The downside of IBP, however, is that it can lead to loose upper bounds.
\citet{mirman2018differentiable, gowal2018effectiveness} propose to combine regular loss and IBP abstract loss as (\ref{eq:prior_obj}).
CROWN-IBP~\citep{zhang2019towards} uses a mixture of linear relaxation and IBP to compute the abstract loss and jointly minimize it with the regular loss. 
While the approaches based on (\ref{eq:prior_obj}) produce state-of-the-art results on a set of benchmarks, this type of works rely on an \textit{ad hoc} scheduler to tune the weights between the regular loss and the abstract loss during training.
In addition, regular loss is a loose lower bound of robust loss and minimizing the regular loss does not directly guide the training to a robust model.
In this paper, we show that it is better to combine adversarial loss and abstract loss while leveraging the efficiency of IBP. 
Moreover, we can
eliminate weight tuning and scheduling in a principled manner.

%% file: tex/methodology.tex
\textbf{Overview.}
Let the perturbed input be $\vx_{\text{adv}}$.
The relations among \textit{adversarial loss}, \textit{robust loss} and \textit{IBP abstract loss} are as follows.
\begin{equation}
    \gL(f_\theta(\vx_{\text{adv}}), y) {\le} {\max_{\vx' {\in} \sS(\vx, \epsilon)}} \gL(f_\theta(\vx'), y) {\le} \gL({-}\underline{\vm}_{\text{IBP}}(\vx, \epsilon); y; \theta)
    \label{eq:relation}
\end{equation}
We note that (\ref{eq:relation}) holds for general adversarial training and provable robustness methods.
Specifically adversarial loss provides a lower bound of robust loss and
minimizing this loss can result in good empirical robustness.
Latent adversarial examples~\citep{balunovic2020adversarial}, for instance, 
can be used to construct a different adversarial loss. 
However, a smaller latent adversarial loss does not necessarily indicate better certified robustness.
COLT~\citep{balunovic2020adversarial} uses multiple regularizers to mitigate this issue. 
On the other hand, minimizing the abstract loss can help to train a network with certified robustness. In this case, the choice of verification methods used in computing the abstract loss can significantly influence the final training outcome. For instance, training with the IBP abstract loss can result in a network that is amenable to IBP verification. The true robustness of the network or the robustness attainable under the given neural network architecture, however, could still be far away from this bound. 
In fact, a small gap between empirical robustness and provable robustness does not necessarily indicate the attainment of good robustness (the extreme case would be a ReLU network with only positive weights). 
Thus, the tightness of both losses relative to \textit{robust loss} is critical to  improving the model's true robustness.

We consider the joint minimization of adversarial loss and abstract loss as a two-objective optimization problem. 
A straightforward way to solve this joint optimization problem is to optimize a weighted sum of the objectives. This leads to the following objective similar to (\ref{eq:prior_obj}):
\begin{equation}
    \gL(\theta) = \kappa_1 \gL(f_\theta(\vx_\text{adv}),y) + \kappa_2 \gL(-\underline{\vm}_\text{IBP};y;\theta)
    \label{eq:adv_obj}
\end{equation}

However, this simple linear-combination formulation is only sensible when
the two objectives are not competing, which is rarely the case.
The conflicting objectives require modeling the trade-off between objectives, and are generally handled by adaptive weight updates~\citep{sener2018multi}.
This approach, however, faces the issue that
even though the stochastic gradients for each objective are unbiased estimates of the corresponding full gradients, the weighted sum of the stochastic gradients is a biased estimate if the weights are associated with the sampled gradients.
This bias can cause instability and local optima issues~\citep{liu2019stochastic}.
In this paper, we leverage moment estimates to compute the weights adaptively and ensures their independence from the corresponding sampled gradients to eliminate the bias. Minimizing the two objectives jointly tightens the approximation of robust loss from both ends.
For efficient training, we develop \toolname using FGSM+random init to
compute \textit{adversarial loss} and IBP to compute \textit{abstract loss}.

\subsection{Joint Training as Two-Objective Optimization}
\begin{figure}[h]
    \centering
    \includegraphics[width=\columnwidth]{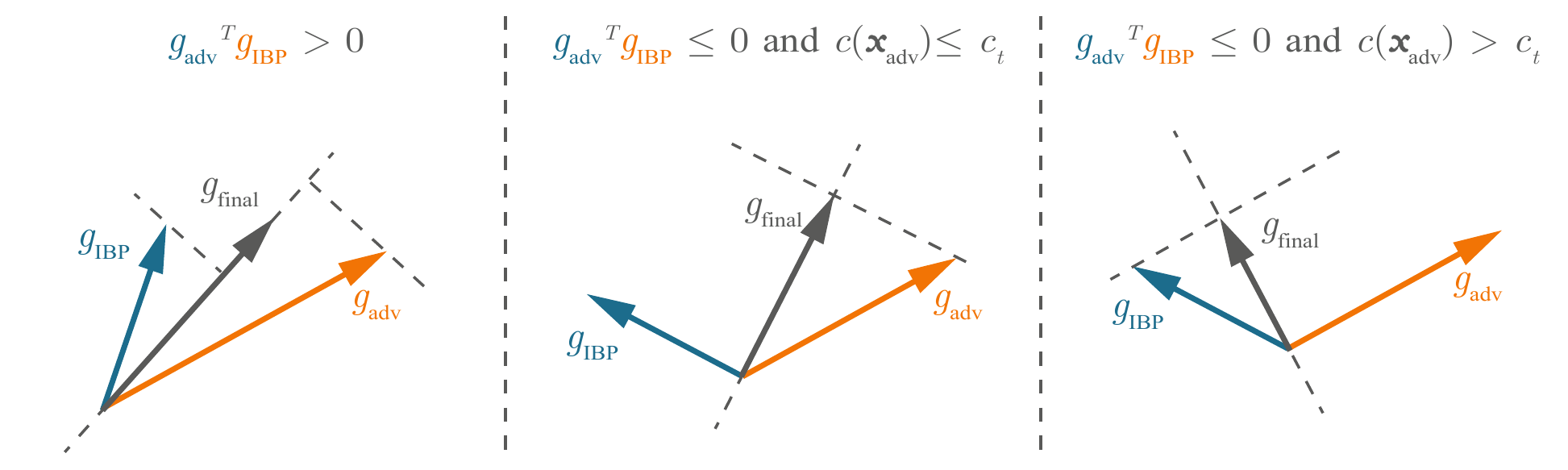}
    \caption{Three cases of computing $g_\text{final}$ from $g_\text{adv}$ and $g_\text{adv}$.}
    \label{fig:gradients}
\end{figure}

We propose a two-objective optimization method inspired by~\citep{fan2019towards, zhang2019learning} to choose the gradient descent direction that reduces \textit{adversarial loss and }\textit{abstract loss} simultaneously.
Let the adversarial loss be $\gL_\text{adv}(\theta)$ and IBP abstract loss be $\overline{\gL}_\text{IBP}(\theta)$. Their gradients with respect to $\theta$ are denoted by
\begin{equation*}
\begin{array}{cc}
     g_{\text{adv}} = \nabla_\theta \gL_\text{adv}(\theta), & g_{\text{IBP}} = \nabla_\theta \overline{\gL}_\text{IBP}(\theta)
\end{array}
\end{equation*}
To balance between the two objectives, we update the network parameters in the direction of the angular bisector of the two gradients. Then, we average the projected vectors of the two gradients on this direction. If $\langle g_{\text{adv}}, g_{\text{IBP}} \rangle {>} 0$, this results in an update that is expected to reduce both losses to improve the adversarial accuracy and tighten IBP.
If $ \langle g_{\text{adv}}, g_{\text{IBP}} \rangle {\le} 0$, taking the angular bisector direction results in an update that improves the objective functions little or not at all for either objective. In this case, we project one of the gradients onto the hyperplane that is perpendicular to the other gradient. The idea is that when two gradients disagree with each other, we prioritize the minimization of one of the objectives. The final gradient guides the search in the direction that reduces the prioritized objective while avoiding increasing the other objective. We use Figure~\ref{fig:gradients} to illustrate this computation.

To decide which direction to prioritize, the tightness of adversarial loss and abstract loss relative to the ground-truth robust loss can be the determining factor. \citet{wang2019convergence} propose the  First-Order Stationary Condition (FOSC) to quantitatively evaluate the adversarial strength of adversarial examples.
In general, the adversarial loss is closer to robust loss with stronger adversarial examples.
Let $c(\vx_\text{adv})$ be FOSC value of $\vx_\text{adv}$ and $c_t$ be the threshold that indicates the desired adversarial strength at the $t$-th epoch. Smaller FOSC values would indicate stronger adversarial examples.
With strong attacks ($c(\vx_\text{adv}) {\le}  c_t$), adversarial training leads to robust models.
Thus, we prioritize the gradient of adversarial loss in this case. The idea is 
to drive the search to the region of robust models with high accuracy and stabilize the minimization of abstract loss.
With weak attacks ($c(\vx_\text{adv}) {>}  c_t$), minimizing adversarial loss does not necessarily imply better robustness.
However, minimizing abstract loss makes solving (\ref{eq: minimax}) tractably.
We prioritize the gradient of abstract loss in this case. \Figref{fig:gradients} provides a visualization of the final gradient computation in different cases.

\begin{algorithm}[hpt]
    \caption{Weight Updates}
    \begin{algorithmic}[1]
        \State \textbf{Input} Exponential decay rates of the moving averages $\beta_1, \beta_2 \in [0, 1)$
        \State \textbf{Init} $\vm_{1, 0} \gets 0$, $\vm_{2, 0} \gets 0$, $\vv_{1, 0} \gets 0$ and $\vv_{2, 0} \gets 0$

        \Procedure{Compute\_weights}{$\vx_\text{adv}, t, c_t$}
            \State $\hat{\vm}_{1, t} \gets \beta_1 \cdot \hat{\vm}_{1, t - 1} + (1 - \beta_1) \cdot g_{\text{adv}, t - 1}$
            \State $\hat{\vm}_{2, t} \gets \beta_1 \cdot \hat{\vm}_{2, t - 1} + (1 - \beta_1) \cdot g_{\text{IBP}, t - 1}$
            \State $\hat{\vv}_{1, t} \gets \beta_2 \cdot \hat{\vv}_{1, t - 1} + (1 - \beta_2) \cdot \Vert g_{\text{adv}, t - 1} \Vert_2$
            \State $\hat{\vv}_{2, t} \gets \beta_2 \cdot \hat{\vv}_{2, t - 1} + (1 - \beta_2) \cdot \Vert g_{\text{IBP}, t - 1} \Vert_2$
            
            \State $\vm_{1, t} {=} \hat{\vm}_{1, t} {/} (1 {-} \beta_1^t)$
            \Comment{Bias-corrected 1st moment}
            \State $\vm_{2, t} {=} \hat{\vm}_{2, t} {/} (1 {-} \beta_1^t)$
            \State $\vv_{1, t} {=} \hat{\vv}_{1, t} {/} (1 {-} \beta_2^t)$
            \Comment{Bias-corrected norm moment}
            \State $\vv_{2, t} {=} \hat{\vv}_{2, t} {/} (1 {-} \beta_2^t)$
            
            \If{$ \langle \vm_{1, t}, \vm_{2, t} \rangle > 0$}
                \State $\gamma {=} \frac{1}{2} \langle \vm_{1, t} {+} \vm_{2, t}, \frac{\vm_{1, t}}{\vv_{1, t}} {+} \frac{\vm_{2, t}}{\vv_{2, t}} \rangle / \Vert \frac{\vm_{1, t}}{\vv_{1, t}} {+} \frac{\vm_{2, t}}{\vv_{2, t}} \Vert_2^2$
                \State $\kappa_\text{adv} = \frac{\gamma}{\vv_{1, t}}$, $\kappa_\text{IBP} = \frac{\gamma}{\vv_{2, t}}$, $\kappa_\text{reg} = 0$
            \Else
                \If{$c(\vx_\text{adv}) \le c_t$} \Comment{check FOSC value}
                    \State $\kappa_\text{adv} = 1$, $\kappa_\text{IBP} = -\frac{\langle \vm_{1, t} \cdot \vm_{2, t} \rangle}{\vv_{2, t}^2}$, $\kappa_\text{reg} = 0$
                \Else
                    \State $\kappa_\text{adv} = -\frac{\langle \vm_{1, t} \cdot \vm_{2, t} \rangle}{\vv_{1, t}^2}$, $\kappa_\text{IBP} = 1, \kappa_\text{reg} = 1/2$
                \EndIf
            \EndIf
            \State \Return{$\kappa_\text{adv}, \kappa_\text{IBP}, \kappa_\text{reg}$}
        \EndProcedure
    \end{algorithmic}
    \label{algorithm:two_obj_training}
\end{algorithm}

\textbf{Stochastic gradients.}
Since the data distribution $\gX$ is unknown in practice, it is impossible to get the full gradients, $g_\text{adv}$ and $g_\text{IBP}$. We denote the realizations of the stochastic objectives at subsequent training epochs $0, \dots, T {-} 1$ as $\gL_{\text{adv}, 0}(\theta^0), {\dots}, \gL_{\text{adv}, T {-} 1}(\theta^{T {-} 1})$ and $\overline{\gL}_{\text{IBP}, 0}(\theta^0), {\dots}, \overline{\gL}_{\text{IBP}, T {-} 1}(\theta^{T {-} 1})$. The stochastic gradients $g_{\text{adv}, t}$ and $g_{\text{IBP}, t}$ are the evaluations of data points from mini-batches and provide unbiased estimation of the full gradients. 
However, the stochastic gradient of the weighted-sum objective at the $t$-th epoch becomes a biased estimate of the final gradient, $g_\text{final}$. The bias is the result of dependence between the weights and the corresponding stochastic gradients.

\textbf{Unbiased weights computation.}
To eliminate this bias, we propose to compute the weights from the estimates of the first and norm moments of the gradients instead of the stochastic gradients. The goal is to ensure the independence of stochastic gradients and the corresponding weights. Let $\vm_{1, t}$, $\vm_{2, t}$, $\vv_{1, t}$ and $\vv_{2, t}$ represent the moment estimates for $g_{\text{adv}, t}$, $g_{\text{IBP}, t}$, $\Vert g_{\text{adv}, t} \Vert_2$ and $\Vert g_{\text{IBP}, t} \Vert_2$ respectively. We modify the moment estimate in~\citep{kingma2014adam} to meet the independence requirement.
In \Algref{algorithm:two_obj_training},
the $t$-th moment estimates are the exponential moving averages of the past stochastic gradients from epoch $0$ to epoch $t {-} 1$, where the hyper-parameters $\beta_1, \beta_2 {\in} [0, 1)$ control the exponential decay rates.
The moving averages themselves, $\vm_{1,t}, \vm_{2, t}, \vv_{1, t}, \vv_{2, t}$, are estimate of the first moment and the norm moment of the true gradients. The independent mini-batch sampling guarantees the independence of stochastic gradients. Thus, the moment estimates are independent from the current sampled stochastic gradient. Then, we calculate the weights using the moment estimates in~\Algref{algorithm:two_obj_training} and update the model parameters with unbiased gradient estimates.

The overall joint training algorithm is shown in \Algref{algorithm:joint_training}.
The regularization term $\kappa_{\text{reg}}$ in line 11 is only used when prioritizing the minimization of abstract loss.
The regularizer helps to bound the convergence rate of training.

\textbf{Leveraging FOSC in joint training.}
In \Algref{algorithm:joint_training}, we use similar dynamic criterion FOSC as in~\citep{wang2019convergence}. In the early stages of training, $c_t$ is close to the maximum FOSC value $c_\text{max}$, which can be satisfied with weak adversarial examples.
Thus, the early stages of training will mostly prioritize the minimization of adversarial loss.
This helps to avoid the instability caused by a loose abstract loss. However, prioritizing the adversarial loss does not necessarily improve the verified robustness of the models.
Thus, we design the FOSC value $c_t$ so that it decreases linearly towards zero as training progresses.
As a result, in the later training stages, the joint training scheme will mostly prioritize the minimization of the abstract loss to improve provable robustness.

\begin{algorithm}[tb]
    \caption{Joint Training}
    \begin{algorithmic}[1]
        \State \textbf{Input} Warm-up epochs $T_\text{nat}$ and $T_\text{adv}$, $\epsilon_\text{train}$ ramp-up epochs $R$,
        maximum FOSC value $c_\text{max}$
        \State $f_{\theta^0} \gets \textsc{Warm-up}(f_{\theta^0}, T_\text{nat}, T_\text{adv})$
        \For{$t = 0$ to $T - 1$}
            \State $c_t {=} \text{clip}(c_\text{max} {-} (t {-} R) {\cdot} c_\text{max} / T' , 0, c_\text{max})$
            \State Sample $\gB {=} \{(\vx_1, y_1), {\dots}, (\vx_\gB, y_\gB)\} {\sim} (\gX, \gY)$
            \For{$i = 0$ to $|\gB| - 1$}
                \State $\epsilon_t {\gets} \textsc{Rampup\_Scheduler}(t, \epsilon_\text{train}, R)$ 
                \State $\vx_{\text{adv}, i} {\gets} \textsc{FGSM+random\_init}(\vx_i, y_i, \epsilon_t)$ 
            \EndFor
            \State $\kappa_\text{adv}, \kappa_\text{IBP}, \kappa_\text{reg} {=} \textsc{Compute\_weights}(\vx_\text{adv}, t, c_t)$ 
            \State $\textit{loss} {=} \kappa_\text{adv} \gL_\text{adv} (\theta^t) {+} \kappa_\text{IBP} \overline{\gL}_\text{IBP} (\theta^t) {+} \kappa_\text{reg} \Vert \overline{\gL}_\text{IBP} (\theta^t) \Vert_2^2$
            \State $\theta^{t + 1} {=} \theta^t {-} \eta_t g_\text{final}(\theta^t)$ \Comment{$g_\text{final}(\theta^t)$: stochastic gradient}
        \EndFor
        \\
        \Procedure{Warmup}{$f_{\theta^0}, T_\text{nat}, T_\text{adv}$} \Comment{Warm-up phase}
            \For{$t =0$ to $T_\text{nat} - 1$}
                \State Train on the \textit{regular loss} $\gL(f_{\theta^t}(\vx), y)$
            \EndFor
            \For{$t = T_\text{nat}$ to $T_\text{nat} + T_\text{adv} - 1$}
                \State Train on the \textit{adversarial loss} $\gL(f_{\theta^t}(\vx_\text{adv}), y)$
            \EndFor
            \State \Return{$f_\theta$}
        \EndProcedure
    \end{algorithmic}
    \label{algorithm:joint_training}
\end{algorithm}

\subsection{Theoretical Analysis}
We provide a theoretical analysis of our proposed joint training scheme to train IBP certified robust networks.
It aims to provide insights on how the ground-truth robust loss changes
during training by our joint training scheme.
The gradient update and the prioritization scheme provide an approximate maximizer for the inner maximization.
Below, we provide theoretical analyses on how \textit{robust loss} changes when two gradients agree with each other and how \textit{abstract loss} changes when two gradients disagree with each other.

In detail, let $\vx^*(\theta) {=} \argmax_{\vx' \in \sS(\vx, \epsilon)} \gL (f_\theta(\vx'), y)$. $\hat{\vx}(\theta)$ is a $\delta$-approximation solution to $\vx^*$, if it satisfies that~\citep{wang2019convergence}
\begin{equation}
    c(\hat{\vx}(\theta)) {=} \max_{\vx' {\in} \sS(\vx, \epsilon)} \langle \vx' {-} \hat{\vx}(\theta), \nabla_{\vx'} \gL(f_\theta(\hat{\vx}(\theta)), y) \rangle {\le} \delta
    \label{eq:delta-optimal}
\end{equation}
Let the robust loss in (\ref{eq: minimax}) be $\gL(\theta)$, and its gradient be $\nabla \gL(\theta) {=} \E [\nabla_\theta \gL(f_\theta(\vx^*(\theta)), y)]$. We denote the stochastic gradient of $\gL(\theta)$ as $g(\theta) = 1 / |\gB| \sum_{i \in \gB} \nabla_\theta \gL(f_\theta(\vx_i^*(\theta)), y_i)$, where $\gB$ is the mini-batch.
Similarly, we denote the abstract loss as $\overline{\gL}(\theta)$, and its gradient as $\nabla \overline{\gL}(\theta) = \E [\nabla_\theta \gL({-}\underline{\vm}(\vx, \epsilon); y; \theta)]$. We denote the stochastic gradient of $\overline{\gL}(\theta)$ as $\overline{g}(\theta) {=} 1 / |\gB| \sum_{i \in \gB} \nabla_\theta \gL({-}\underline{\vm}(\vx_i, \epsilon); y_i; \theta)$. Note that $\E[g(\theta)] {=} \nabla \gL(\theta)$ and $\E[\overline{g}(\theta)] {=} \nabla \overline{\gL}(\theta)$.
The adversarial loss, $\gL_\text{adv}(\theta)$, is $\E [\gL(f_\theta(\hat{\vx}(\theta)), y)]$ and its stochastic gradient is $\hat{g}(\theta) {=} 1/ |\gB| \sum_{i \in \gB} \nabla_\theta \gL(f_\theta(\hat{x}_i(\theta)), y_i)$.
We make assumptions similar to those in \citet{wang2019convergence} and present the theoretical analysis of our method below.

\begin{assumption}
    The function $\gL(\theta;\vx)$ and $\overline{\gL}(\theta;\vx)$ satisfies the gradient Lipschitz conditions s.t.
    \begin{align*}
        \sup_\vx \Vert \nabla_\theta \gL(\theta; \vx) {-} \nabla_\theta \gL(\theta'; \vx) \Vert_2 {\le} L_{\theta \theta} \Vert \theta {-} \theta' \Vert_2 & \text{ } \\ \sup_\vx \Vert \nabla_\theta \overline{\gL}(\theta; \vx) {-} \nabla_\theta \overline{\gL}(\theta'; \vx) \Vert_2 {\le} \overline{L}_{\theta \theta} \Vert \theta {-} \theta' \Vert_2 \\
        \sup_\theta \Vert \nabla_\theta \gL(\theta; \vx) {-} \nabla_\theta \gL(\theta; \vx') \Vert_2 {\le} L_{\theta \vx} \Vert \vx {-} \vx' \Vert_2 & \text{ } \\ \sup_\theta \Vert \nabla_\theta \overline{\gL}(\theta; \vx) {-} \nabla_\theta \overline{\gL}(\theta; \vx') \Vert_2 {\le} \overline{L}_{\theta \vx} \Vert \vx {-} \vx' \Vert_2 \\
        \sup_\vx \Vert \nabla_\vx \gL(\theta; \vx) {-} \nabla_\vx \gL(\theta'; \vx) \Vert_2 {\le} L_{\vx \theta} \Vert \theta {-} \theta' \Vert_2 & \text{ } \\ \sup_\vx \Vert \nabla_\vx \overline{\gL}(\theta; \vx) {-} \nabla_\vx \overline{\gL}(\theta'; \vx) \Vert_2 {\le} \overline{L}_{\vx \theta} \Vert \theta {-} \theta' \Vert_2
    \end{align*}
    where $L_{\theta \theta}$, $L_{\theta \vx}$, $L_{\vx, \theta}, \overline{L}_{\theta \theta}$, $\overline{L}_{\theta \vx}$, $\overline{L}_{\vx, \theta}$ are positive scalars.
    \label{assumption: lipschitz}
\end{assumption}

Assumption~\ref{assumption: lipschitz} was made in~\citep{wang2019convergence} to assume the smoothness of the loss function. Recent studies~\citep{du2019gradient, du2019provablyoptimize} help justify it by showing that the loss function of overparameterized neural networks is semi-smooth.

Let $L {=} (L_{\theta \vx} L_{\vx \theta} / \mu {+} L_{\theta \theta})$ and $\overline{L} {=} (\overline{L}_{\theta \vx} \overline{L}_{\vx \theta} / \mu {+} \overline{L}_{\theta \theta})$. For stochastic gradient descent, we can assume that the variances of stochastic gradients $g(\theta)$ and $\overline{g}(\theta)$ are bounded by constants $\sigma, \overline{\sigma} > 0$.
Let $\Delta{=}\gL(\theta^0) {-} \min_\theta \gL(\theta)$ and $\overline{\Delta}{=}\overline{\gL}(\theta^0) {-} \min_\theta \overline{\gL}(\theta)$.
Under Assumption~\ref{assumption: lipschitz}
, we have the following theoretical results.
\begin{theorem}
    If the dot product of the gradients of the two objectives is greater than 0 and the step size of the training is set to $\eta_t {=} \eta {=} \min(1 / 6 L, \sqrt{\Delta / T L \sigma^2})$, then the expectation of the gradient of \textit{robust loss} satisfies
    \begin{equation*}
        \frac{1}{T} \sum_{t=0}^{T - 1} \E[\Vert \nabla \gL(\theta^t) \Vert_2^2] \le 8 \sigma \sqrt{\frac{L \Delta}{T}} + \frac{7L_{\theta \vx}^2 \delta}{3 \mu}.
    \end{equation*}
    \label{theorem: same_dir}
\end{theorem}

\begin{theorem}
    If the dot product of the gradients of the two objectives is smaller or equal to 0, \textit{adversarial loss} is not tight enough ($c(\vx_\text{adv}) {>} c_t$), and the step size of training is set to $\eta_t {=} \eta {=} \min( 2 * \E[\overline{\gL}_\text{IBP}(\theta^t)] / \overline{L} - 1 / \overline{L}, \sqrt{\overline{\Delta} / T \overline{L} \sigma^2})$ with $\E[\overline{\gL}_\text{IBP}(\theta^t)] {>} 1 / 2$, then the expectation of the gradient of IBP \textit{abstract loss} satisfies
    \begin{equation*}
        \frac{1}{T} \sum_{t=0}^{T - 1} \E[\Vert \nabla \overline{\gL}_\text{IBP}(\theta^t) \Vert_2^2] \le 2 \overline{\sigma} \sqrt{\frac{\overline{L} \overline{\Delta}}{T}} (1 + \sum_{t = 0}^{T - 1} (1 + \E[\overline{\gL}_\text{IBP}(\theta^t)])^2). 
    \end{equation*}
    \label{theorem: diff_dir}
\end{theorem}

The complete proof can be found in the Appendix. If the two gradients agree with each other (i.e. their dot product is greater than 0), Theorem~\ref{theorem: same_dir} suggests that the robust loss minimization can converge to a first-order stationary point at a sublinear rate with sufficiently small $\delta$. Using FOSC ensures that the adversarial loss approximates the \textit{robust loss} up to a precision less than $\delta$ as in (\ref{eq:delta-optimal}). 
Note that it is difficult for the perturbed input $\vx_\text{adv}$ to reach the maximum adversarial strength (minimum FOSC value which is 0) as the model becomes more robust during training. \Algref{algorithm:joint_training} will mostly prioritize the abstract loss minimization when the two gradients disagree with each other since $c_t$ is decreasing to 0. In this case, Theorem~\ref{theorem: diff_dir} suggests that the abstract loss (as obtained by IBP) minimization can converge to a first-order stationary point at a sublinear rate.
Although $\bar{\gL}_{\text{IBP}}$ is not guaranteed to converge, our joint training scheme actively reduces the abstract loss to avoid its divergence.
In practice, potential divergence of the $\bar{\gL}_{\text{IBP}}$ is controlled with a stable training process in our method.
Although Theorem~\ref{theorem: diff_dir} requires $\E [\overline{\gL}_\text{IBP}(\theta^t)] {>} 1/2$, the abstract loss will be sufficiently small if the condition does not hold. With Theorem~\ref{theorem: same_dir} and~\ref{theorem: diff_dir}, the \textit{robust loss} or its upper bound \textit{abstract loss} can be minimized at a sublinear convergence rate. These results provide theoretical support for our approach.

%% file: tex/experiment.tex
\textbf{Experiment setup.}
We evaluate \toolname on all the network model structures used in~\citep{gowal2018effectiveness, zhang2019towards} on the MNIST and CIFAR-10 datasets with different $\evl_\infty$ perturbation bounds, $\epsilon$. We denote these models as \textbf{DM-Small}, \textbf{DM-Medium} and \textbf{DM-Large}.
We perform all experiments on a desktop server using at most 4 GeForce GTX 1080 Ti GPUs.
All models are trained using a single GPU except for \textbf{DM-Large} which requires all 4 GPUs.

\textbf{Metrics.}
We use the following metrics to compare the trained neural networks:
(i) IBP verified error, which is the percentage of test examples that are not verified by IBP, (ii) standard error, which is the test error evaluated on the clean test dataset, and (iii) PGD error, which is the test error under 200-step PGD attack.
Verified errors provide the worst-case test error against $\evl_\infty$ perturbations. PGD errors provide valid lower bounds of test errors against $\evl_\infty$ perturbations.

\begin{figure}[hbp]
    \centering
    \subfloat{\includegraphics[width=0.47\columnwidth]{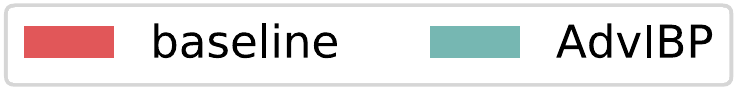}}
    \vfill
    \setcounter{subfigure}{0}
	\subfloat[][MNIST, $\epsilon=0.3$]{
		\includegraphics[width=.45\columnwidth]{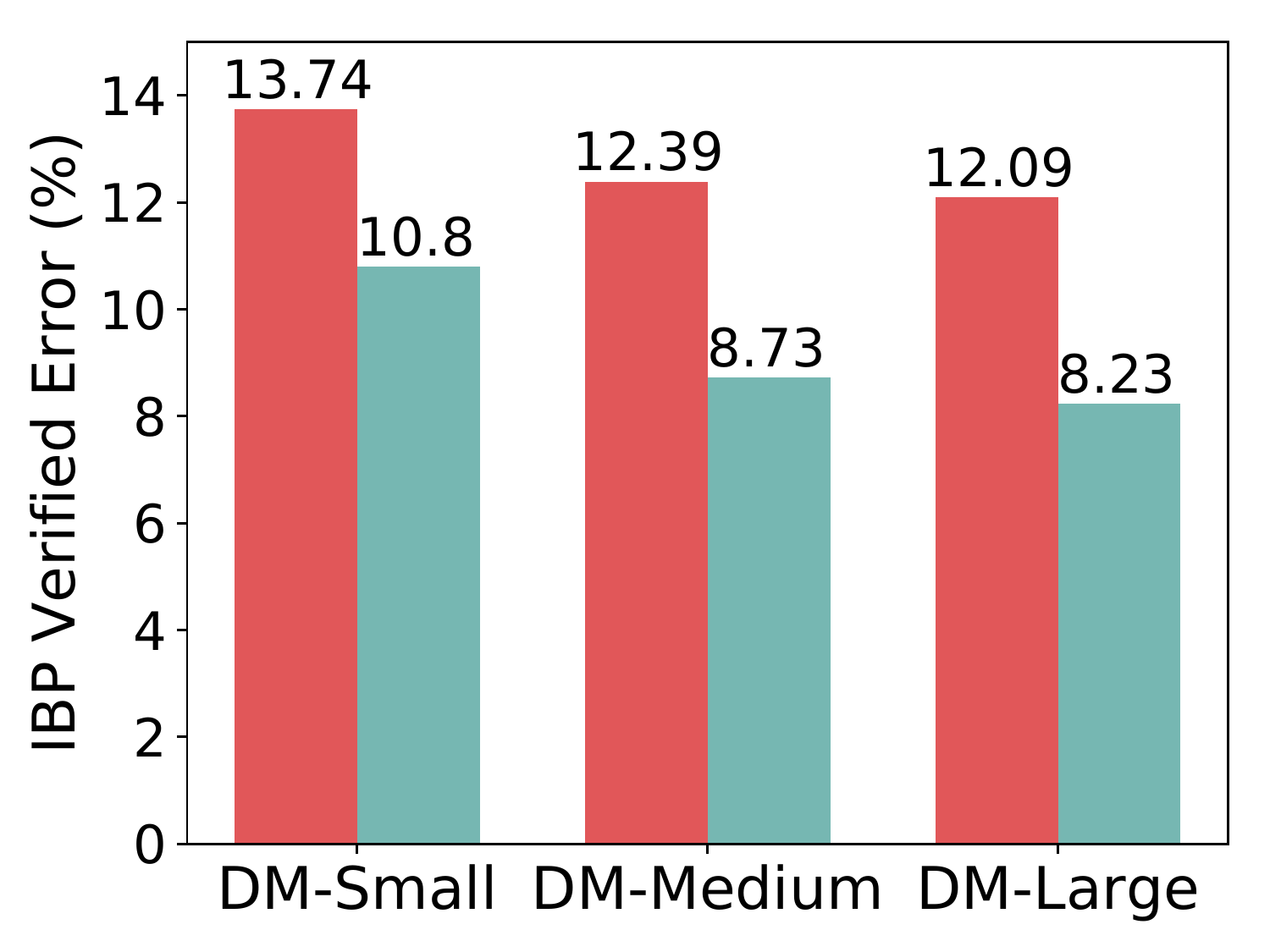}
	}
	\hfill
	\subfloat[][CIFAR-10, $\epsilon=8/255$]{
		\includegraphics[width=.45\columnwidth]{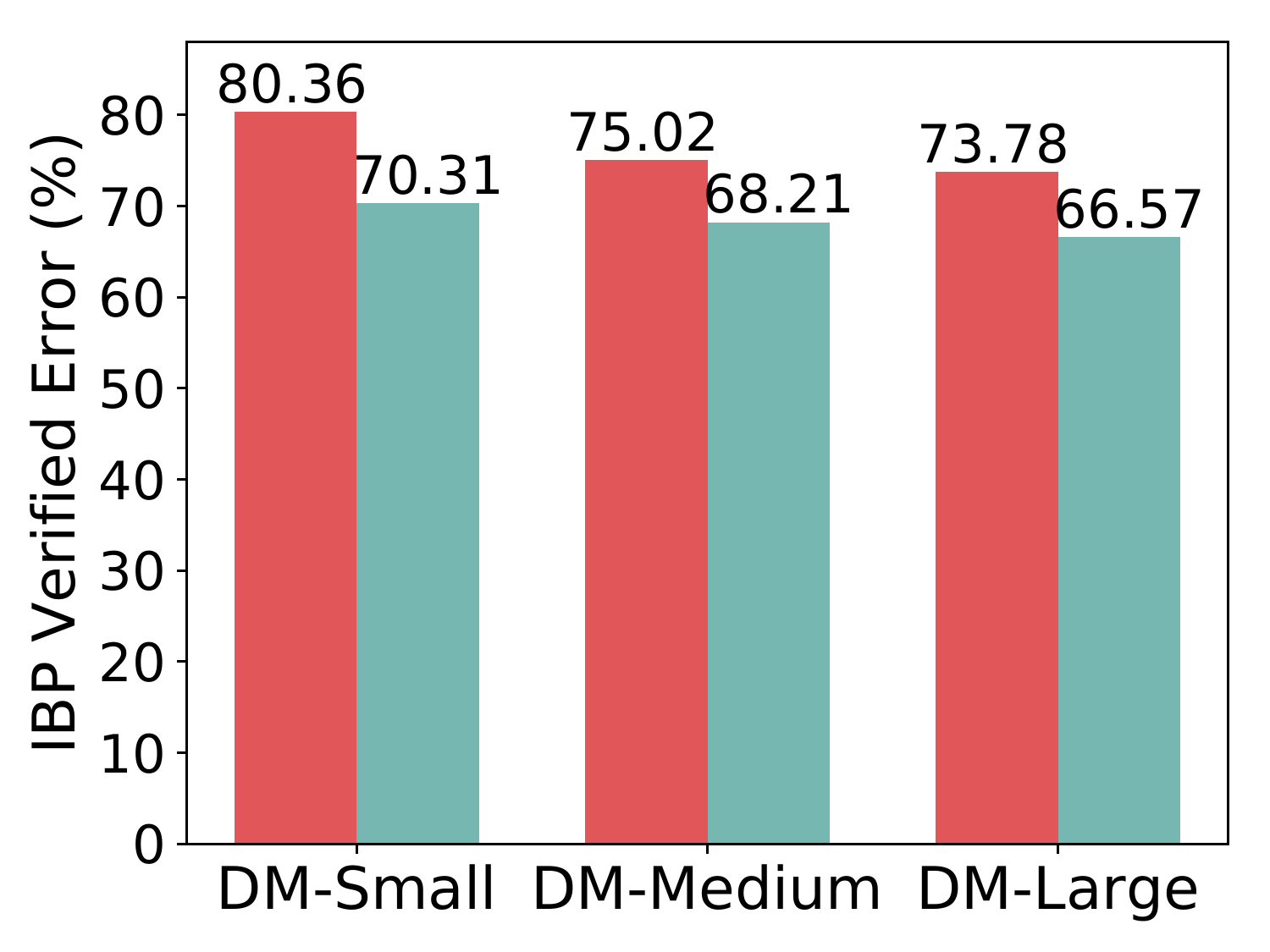}
	}
    \caption{Comparison with the baseline.}
    \label{fig:baseline}
\end{figure}

\textbf{Baseline comparison.}
We consider a baseline method that uses the same warm-up strategy in \Algref{algorithm:joint_training} but fixes the coefficients to $\kappa_{\text{adv}}{=}1.0$ and $\kappa_{\text{IBP}}{=}1.0$ (effectively using the weighted sum method).
As shown in Figure~\ref{fig:baseline}, \toolname, which automatically adapts the coefficients, reduces the IBP verified errors by 9.1\% to 31.9\% compared with the baseline.

\textbf{Comparison with prior works.}
Table~\ref{tab:results} and~\ref{tab:robustness} shows the standard, verified and PGD errors under different $\epsilon$ on CIFAR-10 and MNIST.
On CIFAR-10, our method outperforms the state-of-art methods on verified errors obtained from IBP. In addition to CROWN-IBP, we also present the best errors reported by IBP method~\citep{gowal2018effectiveness}, MIP~\citep{xiao2019training} and COLT~\citep{balunovic2020adversarial}. Note that MIP~\citep{xiao2019training} reports the verified error obtained by mixed integer programming, which is able to compute the \textit{exact} value of \textit{robust loss}. COLT~\citep{balunovic2020adversarial} uses
RefineZono
to compute the verified errors and RefineZono is supposed to a much higher precision than IBP.
On both MNIST and CIFAR-10, even though our method does not use \textit{regular loss}, we still achieve lower standard errors across different models in most cases. The verified errors obtained by \toolname on MNIST can match the prior state-of-art results.
The result of $\evl_\infty$ perturbation $2/255$ outperforms existing approaches except for the results in~\citep{zhang2019towards, singh2019boosting}. However, we note here that both methods in~\citep{zhang2019towards, singh2019boosting} use over-approximation methods with better precision in both training and verification, which may result in significant computation overhead and memory requirement.
We hypothesize that the main reason for this performance gap is that with a relatively small $\evl_\infty$ perturbation,
the minimization of \textit{IBP abstract loss} reduces the capacity of the models 
to learn well as reflected by the higher standard errors.

Additionally, we compare \toolname with CROWN-IBP 
\textit{across a wide range of neural network models}  (Table~\ref{tab:all_models}) rather than on a few hand-selected models. 
In Table~\ref{tab:all_models_results}, we present the best, median and worst verified and standard test errors for models trained on MNIST and CIFAR-10 using CROWN-IBP (with default settings) and \toolname respectively. \toolname's best, median and worst verified errors outperform those of CROWN-IBP in almost all cases.
\begin{table*}[htbp]
\caption{Evaluation on the CIFAR-10 dataset between models trained by \toolname and those by CROWN-IBP. \toolname outperforms the state-of-art, CROWN-IBP, and other best reported results under all perturbation and model settings if IBP is used to compute the verified errors. If different network architectures and more precise verification methods are also considered, our IBP verified errors still outperform the best prior results for both $\epsilon=\frac{8}{255}$ and $\epsilon=\frac{16}{255}$.}
\begin{adjustbox}{width=.98\textwidth,center}
\begin{threeparttable}
\begin{tabular}{cccccccccccccc}
\toprule
                                                                   &                                                  & \multicolumn{3}{c}{}                                                                                                                                                           & \multicolumn{3}{c}{}                                                                                                                                                           & \multicolumn{3}{c}{}                                                                                                                                                           & \multicolumn{3}{c}{}                                                                                                                                                    \\
                                                                   &                                                  & \multicolumn{3}{c}{\multirow{-2}{*}{DM-Small model's err. (\%)}}                                                                                                               & \multicolumn{3}{c}{\multirow{-2}{*}{DM-Medium model's err. (\%)}}                                                                                                              & \multicolumn{3}{c}{\multirow{-2}{*}{DM-Large model's err. (\%)}}                                                                                                               & \multicolumn{3}{c}{\multirow{-2}{*}{Best errors reported in literature (\%)\tnote{2}}}                                                                                           \\ \cline{3-14} 
                                                                   &                                                  & \multicolumn{1}{l|}{}                           & \multicolumn{1}{l|}{}                                                         & \multicolumn{1}{l|}{}                        & \multicolumn{1}{l|}{}                           & \multicolumn{1}{l|}{}                                                         & \multicolumn{1}{l|}{}                        & \multicolumn{1}{l|}{}                           & \multicolumn{1}{l|}{}                                                         & \multicolumn{1}{l|}{}                        & \multicolumn{1}{c|}{}                                            & \multicolumn{1}{c|}{}                           &                          \\
\multirow{-4}{*}{$\epsilon$ ($\evl_\infty$ norm)}                  & \multirow{-4}{*}{Training Method}                & \multicolumn{1}{l|}{\multirow{-2}{*}{Standard}} & \multicolumn{1}{l|}{\multirow{-2}{*}{Verified}}                               & \multicolumn{1}{l|}{\multirow{-2}{*}{PGD}}   & \multicolumn{1}{l|}{\multirow{-2}{*}{Standard}} & \multicolumn{1}{l|}{\multirow{-2}{*}{Verified}}                               & \multicolumn{1}{l|}{\multirow{-2}{*}{PGD}}   & \multicolumn{1}{l|}{\multirow{-2}{*}{Standard}} & \multicolumn{1}{l|}{\multirow{-2}{*}{Verified}}                               & \multicolumn{1}{l|}{\multirow{-2}{*}{PGD}}   & \multicolumn{1}{c|}{\multirow{-2}{*}{Method}}                    & \multicolumn{1}{c|}{\multirow{-2}{*}{Standard}} & \multirow{-2}{*}{Verified} \\ \hline
\multicolumn{1}{c|}{}                                              & \multicolumn{1}{c|}{}                            & \multicolumn{1}{c|}{}                           & \multicolumn{1}{c|}{\cellcolor[HTML]{D5CFCD}}                                 & \multicolumn{1}{c|}{}                        & \multicolumn{1}{c|}{}                           & \multicolumn{1}{c|}{\cellcolor[HTML]{D5CFCD}}                                 & \multicolumn{1}{c|}{}                        & \multicolumn{1}{c|}{}                           & \multicolumn{1}{c|}{\cellcolor[HTML]{D5CFCD}}                                 & \multicolumn{1}{c|}{}                        & \multicolumn{1}{c|}{IBP method~\citep{gowal2018effectiveness}\tnote{3}} & \multicolumn{1}{c|}{39.22}                      & \cellcolor[HTML]{D5CFCD}55.19                      \\
\multicolumn{1}{c|}{}                                              & \multicolumn{1}{c|}{\multirow{-2}{*}{CROWN-IBP}} & \multicolumn{1}{c|}{\multirow{-2}{*}{38.15}}    & \multicolumn{1}{c|}{\multirow{-2}{*}{\cellcolor[HTML]{D5CFCD}52.57}} & \multicolumn{1}{c|}{\multirow{-2}{*}{50.35}} & \multicolumn{1}{c|}{\multirow{-2}{*}{32.78}}    & \multicolumn{1}{c|}{\multirow{-2}{*}{\cellcolor[HTML]{D5CFCD}49.57}} & \multicolumn{1}{c|}{\multirow{-2}{*}{44.22}} & \multicolumn{1}{c|}{\multirow{-2}{*}{28.48}}    & \multicolumn{1}{c|}{\multirow{-2}{*}{\cellcolor[HTML]{D5CFCD}46.03}} & \multicolumn{1}{c|}{\multirow{-2}{*}{40.28}} & \multicolumn{1}{c|}{MIP~\citep{xiao2019training}}              & \multicolumn{1}{c|}{38.88}                      & \cellcolor[HTML]{D5CFCD}54.07                      \\ \cline{2-11}
\multicolumn{1}{c|}{}                                              & \multicolumn{1}{c|}{}                            & \multicolumn{1}{c|}{}                           & \multicolumn{1}{c|}{\cellcolor[HTML]{D5CFCD}}                                 & \multicolumn{1}{c|}{}                        & \multicolumn{1}{c|}{}                           & \multicolumn{1}{c|}{\cellcolor[HTML]{D5CFCD}}                                 & \multicolumn{1}{c|}{}                        & \multicolumn{1}{c|}{}                           & \multicolumn{1}{c|}{\cellcolor[HTML]{D5CFCD}}                                 & \multicolumn{1}{c|}{}                        & \multicolumn{1}{c|}{COLT~\citep{balunovic2020adversarial}}     & \multicolumn{1}{c|}{21.60}                      & \cellcolor[HTML]{D5CFCD}39.50                      \\
\multicolumn{1}{c|}{\multirow{-4}{*}{$\epsilon{=}\frac{2}{255}$\tnote{1}}}  & \multicolumn{1}{c|}{\multirow{-2}{*}{\toolname}}  & \multicolumn{1}{c|}{\multirow{-2}{*}{42.33}}    & \multicolumn{1}{c|}{\multirow{-2}{*}{\cellcolor[HTML]{D5CFCD}56.00}}          & \multicolumn{1}{c|}{\multirow{-2}{*}{50.08}} & \multicolumn{1}{c|}{\multirow{-2}{*}{35.36}}    & \multicolumn{1}{c|}{\multirow{-2}{*}{\cellcolor[HTML]{D5CFCD}52.27}}          & \multicolumn{1}{c|}{\multirow{-2}{*}{43.75}} & \multicolumn{1}{c|}{\multirow{-2}{*}{40.61}}    & \multicolumn{1}{c|}{\multirow{-2}{*}{\cellcolor[HTML]{D5CFCD}51.66}}          & \multicolumn{1}{c|}{\multirow{-2}{*}{46.97}} & \multicolumn{1}{c|}{}                                            & \multicolumn{1}{c|}{}                           & \cellcolor[HTML]{D5CFCD}                           \\ \hline

\multicolumn{1}{c|}{}                                              & \multicolumn{1}{c|}{}                            & \multicolumn{1}{c|}{}                           & \multicolumn{1}{c|}{\cellcolor[HTML]{D5CFCD}}                                 & \multicolumn{1}{c|}{}                        & \multicolumn{1}{c|}{}                           & \multicolumn{1}{c|}{\cellcolor[HTML]{D5CFCD}}                                 & \multicolumn{1}{c|}{}                        & \multicolumn{1}{c|}{}                           & \multicolumn{1}{c|}{\cellcolor[HTML]{D5CFCD}}                                 & \multicolumn{1}{c|}{}                        & \multicolumn{1}{c|}{IBP method~\citep{gowal2018effectiveness}\tnote{3}} & \multicolumn{1}{c|}{58.43}                      & \cellcolor[HTML]{D5CFCD}70.81                      \\
\multicolumn{1}{c|}{}                                              & \multicolumn{1}{c|}{\multirow{-2}{*}{CROWN-IBP}} & \multicolumn{1}{c|}{\multirow{-2}{*}{59.94}}    & \multicolumn{1}{c|}{\multirow{-2}{*}{\cellcolor[HTML]{D5CFCD}70.76}}          & \multicolumn{1}{c|}{\multirow{-2}{*}{69.65}} & \multicolumn{1}{c|}{\multirow{-2}{*}{58.19}}    & \multicolumn{1}{c|}{\multirow{-2}{*}{\cellcolor[HTML]{D5CFCD}68.94}}          & \multicolumn{1}{c|}{\multirow{-2}{*}{67.72}} & \multicolumn{1}{c|}{\multirow{-2}{*}{54.02}}    & \multicolumn{1}{c|}{\multirow{-2}{*}{\cellcolor[HTML]{D5CFCD}66.94}}          & \multicolumn{1}{c|}{\multirow{-2}{*}{65.42}} & \multicolumn{1}{c|}{MIP~\citep{xiao2019training}}              & \multicolumn{1}{c|}{59.55}                      & \cellcolor[HTML]{D5CFCD}79.73                      \\ \cline{2-11}
\multicolumn{1}{c|}{}                                              & \multicolumn{1}{c|}{}                            & \multicolumn{1}{c|}{}                           & \multicolumn{1}{c|}{\cellcolor[HTML]{D5CFCD}}                                 & \multicolumn{1}{c|}{}                        & \multicolumn{1}{c|}{}                           & \multicolumn{1}{c|}{\cellcolor[HTML]{D5CFCD}}                                 & \multicolumn{1}{c|}{}                        & \multicolumn{1}{c|}{}                           & \multicolumn{1}{c|}{\cellcolor[HTML]{D5CFCD}}                                 & \multicolumn{1}{c|}{}                        & \multicolumn{1}{c|}{COLT~\citep{balunovic2020adversarial}}     & \multicolumn{1}{c|}{48.30}                      & \cellcolor[HTML]{D5CFCD}72.50                      \\
\multicolumn{1}{c|}{\multirow{-4}{*}{$\epsilon{=}\frac{8}{255}$}}  & \multicolumn{1}{c|}{\multirow{-2}{*}{\toolname}}  & \multicolumn{1}{c|}{\multirow{-2}{*}{57.88}}    & \multicolumn{1}{c|}{\multirow{-2}{*}{\cellcolor[HTML]{D5CFCD}\textbf{70.31}}} & \multicolumn{1}{c|}{\multirow{-2}{*}{66.52}} & \multicolumn{1}{c|}{\multirow{-2}{*}{54.20}}    & \multicolumn{1}{c|}{\multirow{-2}{*}{\cellcolor[HTML]{D5CFCD}\textbf{68.21}}} & \multicolumn{1}{c|}{\multirow{-2}{*}{61.21}} & \multicolumn{1}{c|}{\multirow{-2}{*}{52.86}}    & \multicolumn{1}{c|}{\multirow{-2}{*}{\cellcolor[HTML]{D5CFCD}\textbf{66.57}}} & \multicolumn{1}{c|}{\multirow{-2}{*}{61.66}} & \multicolumn{1}{c|}{}                                            & \multicolumn{1}{c|}{}                           & \cellcolor[HTML]{D5CFCD}                           \\ \hline
\multicolumn{1}{c|}{}                                              & \multicolumn{1}{c|}{}                            & \multicolumn{1}{c|}{}                           & \multicolumn{1}{c|}{\cellcolor[HTML]{D5CFCD}}                                 & \multicolumn{1}{c|}{}                        & \multicolumn{1}{c|}{}                           & \multicolumn{1}{c|}{\cellcolor[HTML]{D5CFCD}}                                 & \multicolumn{1}{c|}{}                        & \multicolumn{1}{c|}{}                           & \multicolumn{1}{c|}{\cellcolor[HTML]{D5CFCD}}                                 & \multicolumn{1}{c|}{}                        & \multicolumn{1}{c|}{IBP method~\citep{gowal2018effectiveness}\tnote{3}} & \multicolumn{1}{c|}{68.97}                      & \cellcolor[HTML]{D5CFCD}78.12                      \\
\multicolumn{1}{c|}{}                                              & \multicolumn{1}{c|}{\multirow{-2}{*}{CROWN-IBP}} & \multicolumn{1}{c|}{\multirow{-2}{*}{67.42}}    & \multicolumn{1}{c|}{\multirow{-2}{*}{\cellcolor[HTML]{D5CFCD}78.41}}          & \multicolumn{1}{c|}{\multirow{-2}{*}{76.86}} & \multicolumn{1}{c|}{\multirow{-2}{*}{67.94}}    & \multicolumn{1}{c|}{\multirow{-2}{*}{\cellcolor[HTML]{D5CFCD}78.46}}          & \multicolumn{1}{c|}{\multirow{-2}{*}{77.21}} & \multicolumn{1}{c|}{\multirow{-2}{*}{66.06}}    & \multicolumn{1}{c|}{\multirow{-2}{*}{\cellcolor[HTML]{D5CFCD}76.80}}          & \multicolumn{1}{c|}{\multirow{-2}{*}{75.23}} & \multicolumn{1}{c|}{MIP~\citep{xiao2019training}}              & \multicolumn{1}{c|}{n/a}                        & \cellcolor[HTML]{D5CFCD}n/a                         \\ \cline{2-11}
\multicolumn{1}{c|}{}                                              & \multicolumn{1}{c|}{}                            & \multicolumn{1}{c|}{}                           & \multicolumn{1}{c|}{\cellcolor[HTML]{D5CFCD}}                                 & \multicolumn{1}{c|}{}                        & \multicolumn{1}{c|}{}                           & \multicolumn{1}{c|}{\cellcolor[HTML]{D5CFCD}}                                 & \multicolumn{1}{c|}{}                        & \multicolumn{1}{c|}{}                           & \multicolumn{1}{c|}{\cellcolor[HTML]{D5CFCD}}                                 & \multicolumn{1}{c|}{}                        & \multicolumn{1}{c|}{COLT~\citep{balunovic2020adversarial}}     & \multicolumn{1}{c|}{n/a}                        & \cellcolor[HTML]{D5CFCD}n/a                         \\
\multicolumn{1}{c|}{\multirow{-4}{*}{$\epsilon{=}\frac{16}{255}$}} & \multicolumn{1}{c|}{\multirow{-2}{*}{\toolname}}  & \multicolumn{1}{c|}{\multirow{-2}{*}{67.32}}    & \multicolumn{1}{c|}{\multirow{-2}{*}{\cellcolor[HTML]{D5CFCD}\textbf{78.12}}} & \multicolumn{1}{c|}{\multirow{-2}{*}{73.44}} & \multicolumn{1}{c|}{\multirow{-2}{*}{66.26}}    & \multicolumn{1}{c|}{\multirow{-2}{*}{\cellcolor[HTML]{D5CFCD}\textbf{77.79}}} & \multicolumn{1}{c|}{\multirow{-2}{*}{73.52}} & \multicolumn{1}{c|}{\multirow{-2}{*}{64.40}}    & \multicolumn{1}{c|}{\multirow{-2}{*}{\cellcolor[HTML]{D5CFCD}\textbf{76.05}}} & \multicolumn{1}{c|}{\multirow{-2}{*}{71.78}} & \multicolumn{1}{c|}{}                                            & \multicolumn{1}{c|}{}                           & \cellcolor[HTML]{D5CFCD}                           \\ \bottomrule
\end{tabular}
\begin{tablenotes}
    \item[1] The verified error of CROWN-IBP in this setting is computed using CROWN.
    \item[2] Some of the best errors from literature are obtained from models with different architectures from ours. Some of the verified errors are also obtained using more precise verification methods.
    \item[3] The results are reproduced by~\citep{zhang2019towards} on the same perturbation settings and models used by our method and CROWN-IBP. The verified error is obtained from IBP.
\end{tablenotes}
\end{threeparttable}
\end{adjustbox}
\label{tab:results}
\end{table*}

\begin{table*}[htpb]
\caption{Standard, verified and PGD test errors \textit{for a wide range of models} trained on MNIST and CIFAR-10 datasets using CROWN-IBP and \toolname. The purpose of this experiment is to compare model performance statistics on a wide range of models, rather than a few selected models. For each settings, we report 3 statistics, the smallest, median and largest verified errors. We also report the standard and PGD errors in the same way.}
\begin{adjustbox}{width=0.8\textwidth,center}
\begin{threeparttable}
\begin{tabular}{ccccccccccccc}
\toprule
 &  &  & \multicolumn{3}{c}{} & \multicolumn{3}{c}{} & \multicolumn{3}{c}{} &  \\
 &  &  & \multicolumn{3}{c}{\multirow{-2}{*}{Standard Error (\%)}} & \multicolumn{3}{c}{\multirow{-2}{*}{Verified Error (\%)}} & \multicolumn{3}{c}{\multirow{-2}{*}{PGD Error (\%)}} &  \\ \cline{4-12}
 &  &  &  &  & \multicolumn{1}{c|}{} & \cellcolor[HTML]{D5CFCD} & \cellcolor[HTML]{D5CFCD} & \multicolumn{1}{c|}{\cellcolor[HTML]{D5CFCD}} &  &  &  &  \\
\multirow{-4}{*}{Dataset} & \multirow{-4}{*}{$\epsilon$ ($\evl_\infty$ norm)} & \multirow{-4}{*}{Training Method} & \multirow{-2}{*}{best} & \multirow{-2}{*}{median} & \multicolumn{1}{c|}{\multirow{-2}{*}{worst}} & \multirow{-2}{*}{\cellcolor[HTML]{D5CFCD}best} & \multirow{-2}{*}{\cellcolor[HTML]{D5CFCD}median} & \multicolumn{1}{c|}{\multirow{-2}{*}{\cellcolor[HTML]{D5CFCD}worst}} & \multirow{-2}{*}{best} & \multirow{-2}{*}{median} & \multirow{-2}{*}{worst} & \multirow{-4}{*}{\begin{tabular}[c]{@{}c@{}}Number of AdvIBP models\\ with lower verified errors\\ among all trained model structures\end{tabular}} \\ \hline
\multicolumn{1}{c|}{} & \multicolumn{1}{c|}{} & \multicolumn{1}{c|}{} & \multicolumn{1}{c|}{} & \multicolumn{1}{c|}{} & \multicolumn{1}{c|}{} & \multicolumn{1}{c|}{\cellcolor[HTML]{D5CFCD}} & \multicolumn{1}{c|}{\cellcolor[HTML]{D5CFCD}} & \multicolumn{1}{c|}{\cellcolor[HTML]{D5CFCD}} & \multicolumn{1}{c|}{} & \multicolumn{1}{c|}{} & \multicolumn{1}{c|}{} &  \\
\multicolumn{1}{c|}{} & \multicolumn{1}{c|}{} & \multicolumn{1}{c|}{\multirow{-2}{*}{CROWN-IBP}} & \multicolumn{1}{c|}{\multirow{-2}{*}{2.49}} & \multicolumn{1}{c|}{\multirow{-2}{*}{3.50}} & \multicolumn{1}{c|}{\multirow{-2}{*}{5.39}} & \multicolumn{1}{c|}{\multirow{-2}{*}{\cellcolor[HTML]{D5CFCD}4.81}} & \multicolumn{1}{c|}{\multirow{-2}{*}{\cellcolor[HTML]{D5CFCD}6.33}} & \multicolumn{1}{c|}{\multirow{-2}{*}{\cellcolor[HTML]{D5CFCD}8.82}} & \multicolumn{1}{c|}{\multirow{-2}{*}{3.42}} & \multicolumn{1}{c|}{\multirow{-2}{*}{4.94}} & \multicolumn{1}{c|}{\multirow{-2}{*}{7.33}} &  \\ \cline{3-12}
\multicolumn{1}{c|}{} & \multicolumn{1}{c|}{} & \multicolumn{1}{c|}{} & \multicolumn{1}{c|}{} & \multicolumn{1}{c|}{} & \multicolumn{1}{c|}{} & \multicolumn{1}{c|}{\cellcolor[HTML]{D5CFCD}} & \multicolumn{1}{c|}{\cellcolor[HTML]{D5CFCD}} & \multicolumn{1}{c|}{\cellcolor[HTML]{D5CFCD}} & \multicolumn{1}{c|}{} & \multicolumn{1}{c|}{} & \multicolumn{1}{c|}{} &  \\
\multicolumn{1}{c|}{} & \multicolumn{1}{c|}{\multirow{-4}{*}{$\epsilon=0.2$}} & \multicolumn{1}{c|}{\multirow{-2}{*}{AdvIBP}} & \multicolumn{1}{c|}{\multirow{-2}{*}{\textbf{2.41}}} & \multicolumn{1}{c|}{\multirow{-2}{*}{\textbf{3.36}}} & \multicolumn{1}{c|}{\multirow{-2}{*}{\textbf{5.29}}} & \multicolumn{1}{c|}{\multirow{-2}{*}{\cellcolor[HTML]{D5CFCD}\textbf{4.76}}} & \multicolumn{1}{c|}{\multirow{-2}{*}{\cellcolor[HTML]{D5CFCD}\textbf{6.13}}} & \multicolumn{1}{c|}{\multirow{-2}{*}{\cellcolor[HTML]{D5CFCD}\textbf{8.52}}} & \multicolumn{1}{c|}{\multirow{-2}{*}{\textbf{3.31}}} & \multicolumn{1}{c|}{\multirow{-2}{*}{\textbf{4.70}}} & \multicolumn{1}{c|}{\multirow{-2}{*}{\textbf{7.01}}} & \multirow{-4}{*}{9/10} \\ \cline{2-13} 
\multicolumn{1}{c|}{} & \multicolumn{1}{c|}{} & \multicolumn{1}{c|}{} & \multicolumn{1}{c|}{} & \multicolumn{1}{c|}{} & \multicolumn{1}{c|}{} & \multicolumn{1}{c|}{\cellcolor[HTML]{D5CFCD}} & \multicolumn{1}{c|}{\cellcolor[HTML]{D5CFCD}} & \multicolumn{1}{c|}{\cellcolor[HTML]{D5CFCD}} & \multicolumn{1}{c|}{} & \multicolumn{1}{c|}{} & \multicolumn{1}{c|}{} &  \\
\multicolumn{1}{c|}{} & \multicolumn{1}{c|}{} & \multicolumn{1}{c|}{\multirow{-2}{*}{CROWN-IBP}} & \multicolumn{1}{c|}{\multirow{-2}{*}{2.49}} & \multicolumn{1}{c|}{\multirow{-2}{*}{3.50}} & \multicolumn{1}{c|}{\multirow{-2}{*}{5.39}} & \multicolumn{1}{c|}{\multirow{-2}{*}{\cellcolor[HTML]{D5CFCD}\textbf{7.19}}} & \multicolumn{1}{c|}{\multirow{-2}{*}{\cellcolor[HTML]{D5CFCD}9.12}} & \multicolumn{1}{c|}{\multirow{-2}{*}{\cellcolor[HTML]{D5CFCD}11.58}} & \multicolumn{1}{c|}{\multirow{-2}{*}{3.85}} & \multicolumn{1}{c|}{\multirow{-2}{*}{5.47}} & \multicolumn{1}{c|}{\multirow{-2}{*}{8.46}} &  \\ \cline{3-12}
\multicolumn{1}{c|}{} & \multicolumn{1}{c|}{} & \multicolumn{1}{c|}{} & \multicolumn{1}{c|}{} & \multicolumn{1}{c|}{} & \multicolumn{1}{c|}{} & \multicolumn{1}{c|}{\cellcolor[HTML]{D5CFCD}} & \multicolumn{1}{c|}{\cellcolor[HTML]{D5CFCD}} & \multicolumn{1}{c|}{\cellcolor[HTML]{D5CFCD}} & \multicolumn{1}{c|}{} & \multicolumn{1}{c|}{} & \multicolumn{1}{c|}{} &  \\
\multicolumn{1}{c|}{\multirow{-8}{*}{MNIST}} & \multicolumn{1}{c|}{\multirow{-4}{*}{$\epsilon=0.3$}} & \multicolumn{1}{c|}{\multirow{-2}{*}{AdvIBP}} & \multicolumn{1}{c|}{\multirow{-2}{*}{\textbf{2.41}}} & \multicolumn{1}{c|}{\multirow{-2}{*}{\textbf{3.36}}} & \multicolumn{1}{c|}{\multirow{-2}{*}{\textbf{5.29}}} & \multicolumn{1}{c|}{\multirow{-2}{*}{\cellcolor[HTML]{D5CFCD}7.21}} & \multicolumn{1}{c|}{\multirow{-2}{*}{\cellcolor[HTML]{D5CFCD}\textbf{8.86}}} & \multicolumn{1}{c|}{\multirow{-2}{*}{\cellcolor[HTML]{D5CFCD}\textbf{11.32}}} & \multicolumn{1}{c|}{\multirow{-2}{*}{\textbf{4.04}}} & \multicolumn{1}{c|}{\multirow{-2}{*}{\textbf{5.40}}} & \multicolumn{1}{c|}{\multirow{-2}{*}{\textbf{8.00}}} & \multirow{-4}{*}{8/10} \\ \hline
\multicolumn{1}{c|}{} & \multicolumn{1}{c|}{} & \multicolumn{1}{c|}{} & \multicolumn{1}{c|}{} & \multicolumn{1}{c|}{} & \multicolumn{1}{c|}{} & \multicolumn{1}{c|}{\cellcolor[HTML]{D5CFCD}} & \multicolumn{1}{c|}{\cellcolor[HTML]{D5CFCD}} & \multicolumn{1}{c|}{\cellcolor[HTML]{D5CFCD}} & \multicolumn{1}{c|}{} & \multicolumn{1}{c|}{} & \multicolumn{1}{c|}{} &  \\
\multicolumn{1}{c|}{} & \multicolumn{1}{c|}{} & \multicolumn{1}{c|}{\multirow{-2}{*}{CROWN-IBP}} & \multicolumn{1}{c|}{\multirow{-2}{*}{57.25}} & \multicolumn{1}{c|}{\multirow{-2}{*}{59.84}} & \multicolumn{1}{c|}{\multirow{-2}{*}{63.46}} & \multicolumn{1}{c|}{\multirow{-2}{*}{\cellcolor[HTML]{D5CFCD}69.02}} & \multicolumn{1}{c|}{\multirow{-2}{*}{\cellcolor[HTML]{D5CFCD}71.32}} & \multicolumn{1}{c|}{\multirow{-2}{*}{\cellcolor[HTML]{D5CFCD}72.40}} & \multicolumn{1}{c|}{\multirow{-2}{*}{65.56}} & \multicolumn{1}{c|}{\multirow{-2}{*}{67.57}} & \multicolumn{1}{c|}{\multirow{-2}{*}{70.17}} &  \\ \cline{3-12}
\multicolumn{1}{c|}{} & \multicolumn{1}{c|}{} & \multicolumn{1}{c|}{} & \multicolumn{1}{c|}{} & \multicolumn{1}{c|}{} & \multicolumn{1}{c|}{} & \multicolumn{1}{c|}{\cellcolor[HTML]{D5CFCD}} & \multicolumn{1}{c|}{\cellcolor[HTML]{D5CFCD}} & \multicolumn{1}{c|}{\cellcolor[HTML]{D5CFCD}} & \multicolumn{1}{c|}{} & \multicolumn{1}{c|}{} & \multicolumn{1}{c|}{} &  \\
\multicolumn{1}{c|}{\multirow{-4}{*}{CIFAR-10}} & \multicolumn{1}{c|}{\multirow{-4}{*}{$\epsilon=\frac{8}{255}$}} & \multicolumn{1}{c|}{\multirow{-2}{*}{AdvIBP}} & \multicolumn{1}{c|}{\multirow{-2}{*}{\textbf{57.03}}} & \multicolumn{1}{c|}{\multirow{-2}{*}{\textbf{58.85}}} & \multicolumn{1}{c|}{\multirow{-2}{*}{\textbf{60.97}}} & \multicolumn{1}{c|}{\multirow{-2}{*}{\cellcolor[HTML]{D5CFCD}\textbf{68.50}}} & \multicolumn{1}{c|}{\multirow{-2}{*}{\cellcolor[HTML]{D5CFCD}\textbf{69.36}}} & \multicolumn{1}{c|}{\multirow{-2}{*}{\cellcolor[HTML]{D5CFCD}\textbf{71.40}}} & \multicolumn{1}{c|}{\multirow{-2}{*}{\textbf{65.08}}} & \multicolumn{1}{c|}{\multirow{-2}{*}{\textbf{66.90}}} & \multicolumn{1}{c|}{\multirow{-2}{*}{\textbf{68.74}}} & \multirow{-4}{*}{7/7} \\ \bottomrule
\end{tabular}
\end{threeparttable}
\end{adjustbox}
\label{tab:all_models_results}
\end{table*}

\textbf{\crowntoolname.} 
In our joint training scheme, one can replace IBP with a more precise method for computing the abstract loss. 
We present here the results of
\crowntoolname 
which uses CROWN-IBP to compute the abstract loss on the MNIST dataset.
CROWN-IBP uses a linear combination of CROWN bounds and IBP bounds to compute the abstract loss during the warm-up period.
After the warm-up period, the abstract loss is computed solely with IBP bounds.
In Table~\ref{tab:robustness}, we can observe that with a more precise abstract loss, our joint training scheme outperforms CROWN-IBP and \toolname in IBP verified errors consistently across different model structures. 
In fact, to the best of our knowledge,  \crowntoolname achieves 
the \textit{best} verified error rates  
compared to those reported in existing literature on the MNIST dataset across different choices of network models for these $\epsilon$ bounds.

\begin{table*}[htpb]
\caption[Caption for LOF]{Evaluation on the MNIST dataset between models trained by \toolname, \crowntoolname and those by CROWN-IBP.
The CROWN-IBP result is from Table C. in~\citep{zhang2019towards}. \toolname achieves competitive performance compared to CROWN-IBP on MNIST.
\crowntoolname outperforms CROWN-IBP under all
settings, and achieves state-of-the-art verified errors on MNIST dataset for $l_\infty$ robustness.}
\begin{adjustbox}{width=0.63\textwidth,center}
\begin{threeparttable}
\begin{tabular}{ccccccccccc}
\toprule
 &  & \multicolumn{3}{c}{} & \multicolumn{3}{c}{} & \multicolumn{3}{c}{} \\
 &  & \multicolumn{3}{c}{\multirow{-2}{*}{DM-Small model's err. (\%)}} & \multicolumn{3}{c}{\multirow{-2}{*}{DM-Medium model's err. (\%)}} & \multicolumn{3}{c}{\multirow{-2}{*}{DM-Large model's err. (\%)}} \\ \cline{3-11} 
 &  & \multicolumn{1}{l|}{} & \multicolumn{1}{c|}{\cellcolor[HTML]{D5CFCD}} & \multicolumn{1}{c|}{} & \multicolumn{1}{c|}{} & \multicolumn{1}{c|}{\cellcolor[HTML]{D5CFCD}} & \multicolumn{1}{c|}{} & \multicolumn{1}{c|}{} & \multicolumn{1}{c|}{\cellcolor[HTML]{D5CFCD}} &  \\
\multirow{-4}{*}{$\epsilon$ ($\evl_\infty$ norm)} & \multirow{-4}{*}{Training Method} & \multicolumn{1}{l|}{\multirow{-2}{*}{Standard}} & \multicolumn{1}{c|}{\multirow{-2}{*}{\cellcolor[HTML]{D5CFCD}Verified\tnote{1}}} & \multicolumn{1}{c|}{\multirow{-2}{*}{PGD}} & \multicolumn{1}{c|}{\multirow{-2}{*}{Standard}} & \multicolumn{1}{c|}{\multirow{-2}{*}{\cellcolor[HTML]{D5CFCD}Verified\tnote{1}}} & \multicolumn{1}{c|}{\multirow{-2}{*}{PGD}} & \multicolumn{1}{c|}{\multirow{-2}{*}{Standard}} & \multicolumn{1}{c|}{\multirow{-2}{*}{\cellcolor[HTML]{D5CFCD}Verified\tnote{1}}} & \multirow{-2}{*}{PGD} \\ \hline
\multicolumn{1}{c|}{} & \multicolumn{1}{c|}{} & \multicolumn{1}{c|}{} & \multicolumn{1}{c|}{\cellcolor[HTML]{D5CFCD}} & \multicolumn{1}{c|}{} & \multicolumn{1}{c|}{} & \multicolumn{1}{c|}{\cellcolor[HTML]{D5CFCD}} & \multicolumn{1}{c|}{} & \multicolumn{1}{c|}{} & \multicolumn{1}{c|}{\cellcolor[HTML]{D5CFCD}} &  \\
\multicolumn{1}{c|}{} & \multicolumn{1}{c|}{\multirow{-2}{*}{CROWN-IBP}} & \multicolumn{1}{c|}{\multirow{-2}{*}{1.67}} & \multicolumn{1}{c|}{\multirow{-2}{*}{\cellcolor[HTML]{D5CFCD}3.44}} & \multicolumn{1}{c|}{\multirow{-2}{*}{3.09}} & \multicolumn{1}{c|}{\multirow{-2}{*}{1.14}} & \multicolumn{1}{c|}{\multirow{-2}{*}{\cellcolor[HTML]{D5CFCD}\textbf{2.64}}} & \multicolumn{1}{c|}{\multirow{-2}{*}{2.23}} & \multicolumn{1}{c|}{\multirow{-2}{*}{0.97}} & \multicolumn{1}{c|}{\multirow{-2}{*}{\cellcolor[HTML]{D5CFCD}2.25}} & \multirow{-2}{*}{1.81} \\ \cline{2-11} 
\multicolumn{1}{c|}{} & \multicolumn{1}{c|}{} & \multicolumn{1}{c|}{} & \multicolumn{1}{c|}{\cellcolor[HTML]{D5CFCD}} & \multicolumn{1}{c|}{} & \multicolumn{1}{c|}{} & \multicolumn{1}{c|}{\cellcolor[HTML]{D5CFCD}} & \multicolumn{1}{c|}{} & \multicolumn{1}{c|}{} & \multicolumn{1}{c|}{\cellcolor[HTML]{D5CFCD}} &  \\
\multicolumn{1}{c|}{} & \multicolumn{1}{c|}{\multirow{-2}{*}{AdvIBP\tnote{2}}} & \multicolumn{1}{c|}{\multirow{-2}{*}{1.63}} & \multicolumn{1}{c|}{\multirow{-2}{*}{\cellcolor[HTML]{D5CFCD}3.69}} & \multicolumn{1}{c|}{\multirow{-2}{*}{2.70}} & \multicolumn{1}{c|}{\multirow{-2}{*}{1.41}} & \multicolumn{1}{c|}{\multirow{-2}{*}{\cellcolor[HTML]{D5CFCD}3.24}} & \multicolumn{1}{c|}{\multirow{-2}{*}{2.26}} & \multicolumn{1}{c|}{\multirow{-2}{*}{1.03}} & \multicolumn{1}{c|}{\multirow{-2}{*}{\cellcolor[HTML]{D5CFCD}2.28}} & \multirow{-2}{*}{1.53} \\ \cline{2-11} 
\multicolumn{1}{c|}{} & \multicolumn{1}{c|}{} & \multicolumn{1}{c|}{} & \multicolumn{1}{c|}{\cellcolor[HTML]{D5CFCD}} & \multicolumn{1}{c|}{} & \multicolumn{1}{c|}{} & \multicolumn{1}{c|}{\cellcolor[HTML]{D5CFCD}} & \multicolumn{1}{c|}{} & \multicolumn{1}{c|}{} & \multicolumn{1}{c|}{\cellcolor[HTML]{D5CFCD}} &  \\
\multicolumn{1}{c|}{\multirow{-6}{*}{$\epsilon=0.1$}} & \multicolumn{1}{c|}{\multirow{-2}{*}{AdvCROWN-IBP}} & \multicolumn{1}{c|}{\multirow{-2}{*}{1.52}} & \multicolumn{1}{c|}{\multirow{-2}{*}{\cellcolor[HTML]{D5CFCD}\textbf{3.19}}} & \multicolumn{1}{c|}{\multirow{-2}{*}{2.39}} & \multicolumn{1}{c|}{\multirow{-2}{*}{1.23}} & \multicolumn{1}{c|}{\multirow{-2}{*}{\cellcolor[HTML]{D5CFCD}2.88}} & \multicolumn{1}{c|}{\multirow{-2}{*}{2.18}} & \multicolumn{1}{c|}{\multirow{-2}{*}{1.22}} & \multicolumn{1}{c|}{\multirow{-2}{*}{\cellcolor[HTML]{D5CFCD}\textbf{2.19}}} & \multirow{-2}{*}{1.57} \\ \hline
\multicolumn{1}{c|}{} & \multicolumn{1}{c|}{} & \multicolumn{1}{c|}{} & \multicolumn{1}{c|}{\cellcolor[HTML]{D5CFCD}} & \multicolumn{1}{c|}{} & \multicolumn{1}{c|}{} & \multicolumn{1}{c|}{\cellcolor[HTML]{D5CFCD}} & \multicolumn{1}{c|}{} & \multicolumn{1}{c|}{} & \multicolumn{1}{c|}{\cellcolor[HTML]{D5CFCD}} &  \\
\multicolumn{1}{c|}{} & \multicolumn{1}{c|}{\multirow{-2}{*}{CROWN-IBP}} & \multicolumn{1}{c|}{\multirow{-2}{*}{2.96}} & \multicolumn{1}{c|}{\multirow{-2}{*}{\cellcolor[HTML]{D5CFCD}6.11}} & \multicolumn{1}{c|}{\multirow{-2}{*}{5.74}} & \multicolumn{1}{c|}{\multirow{-2}{*}{2.37}} & \multicolumn{1}{c|}{\multirow{-2}{*}{\cellcolor[HTML]{D5CFCD}5.35}} & \multicolumn{1}{c|}{\multirow{-2}{*}{4.90}} & \multicolumn{1}{c|}{\multirow{-2}{*}{1.62}} & \multicolumn{1}{c|}{\multirow{-2}{*}{\cellcolor[HTML]{D5CFCD}3.87}} & \multirow{-2}{*}{3.81} \\ \cline{2-11} 
\multicolumn{1}{c|}{} & \multicolumn{1}{c|}{} & \multicolumn{1}{c|}{} & \multicolumn{1}{c|}{\cellcolor[HTML]{D5CFCD}} & \multicolumn{1}{c|}{} & \multicolumn{1}{c|}{} & \multicolumn{1}{c|}{\cellcolor[HTML]{D5CFCD}} & \multicolumn{1}{c|}{} & \multicolumn{1}{c|}{} & \multicolumn{1}{c|}{\cellcolor[HTML]{D5CFCD}} &  \\
\multicolumn{1}{c|}{} & \multicolumn{1}{c|}{\multirow{-2}{*}{AdvIBP\tnote{2}}} & \multicolumn{1}{c|}{\multirow{-2}{*}{4.15}} & \multicolumn{1}{c|}{\multirow{-2}{*}{\cellcolor[HTML]{D5CFCD}7.68}} & \multicolumn{1}{c|}{\multirow{-2}{*}{5.81}} & \multicolumn{1}{c|}{\multirow{-2}{*}{2.33}} & \multicolumn{1}{c|}{\multirow{-2}{*}{\cellcolor[HTML]{D5CFCD}5.37}} & \multicolumn{1}{c|}{\multirow{-2}{*}{3.54}} & \multicolumn{1}{c|}{\multirow{-2}{*}{1.58}} & \multicolumn{1}{c|}{\multirow{-2}{*}{\cellcolor[HTML]{D5CFCD}4.70}} & \multirow{-2}{*}{2.59} \\ \cline{2-11} 
\multicolumn{1}{c|}{} & \multicolumn{1}{c|}{} & \multicolumn{1}{c|}{} & \multicolumn{1}{c|}{\cellcolor[HTML]{D5CFCD}} & \multicolumn{1}{c|}{} & \multicolumn{1}{c|}{} & \multicolumn{1}{c|}{\cellcolor[HTML]{D5CFCD}} & \multicolumn{1}{c|}{} & \multicolumn{1}{c|}{} & \multicolumn{1}{c|}{\cellcolor[HTML]{D5CFCD}} &  \\
\multicolumn{1}{c|}{\multirow{-6}{*}{$\epsilon=0.2$}} & \multicolumn{1}{c|}{\multirow{-2}{*}{AdvCROWN-IBP}} & \multicolumn{1}{c|}{\multirow{-2}{*}{3.22}} & \multicolumn{1}{c|}{\multirow{-2}{*}{\cellcolor[HTML]{D5CFCD}\textbf{6.02}}} & \multicolumn{1}{c|}{\multirow{-2}{*}{4.50}} & \multicolumn{1}{c|}{\multirow{-2}{*}{2.45}} & \multicolumn{1}{c|}{\multirow{-2}{*}{\cellcolor[HTML]{D5CFCD}\textbf{5.16}}} & \multicolumn{1}{c|}{\multirow{-2}{*}{3.27}} & \multicolumn{1}{c|}{\multirow{-2}{*}{1.51}} & \multicolumn{1}{c|}{\multirow{-2}{*}{\cellcolor[HTML]{D5CFCD}\textbf{3.87}}} & \multirow{-2}{*}{1.98} \\ \hline
\multicolumn{1}{c|}{} & \multicolumn{1}{c|}{} & \multicolumn{1}{c|}{} & \multicolumn{1}{c|}{\cellcolor[HTML]{D5CFCD}} & \multicolumn{1}{c|}{} & \multicolumn{1}{c|}{} & \multicolumn{1}{c|}{\cellcolor[HTML]{D5CFCD}} & \multicolumn{1}{c|}{} & \multicolumn{1}{c|}{} & \multicolumn{1}{c|}{\cellcolor[HTML]{D5CFCD}} &  \\
\multicolumn{1}{c|}{} & \multicolumn{1}{c|}{\multirow{-2}{*}{CROWN-IBP}} & \multicolumn{1}{c|}{\multirow{-2}{*}{3.55}} & \multicolumn{1}{c|}{\multirow{-2}{*}{\cellcolor[HTML]{D5CFCD}9.40}} & \multicolumn{1}{c|}{\multirow{-2}{*}{8.50}} & \multicolumn{1}{c|}{\multirow{-2}{*}{2.37}} & \multicolumn{1}{c|}{\multirow{-2}{*}{\cellcolor[HTML]{D5CFCD}8.54}} & \multicolumn{1}{c|}{\multirow{-2}{*}{7.74}} & \multicolumn{1}{c|}{\multirow{-2}{*}{1.62}} & \multicolumn{1}{c|}{\multirow{-2}{*}{\cellcolor[HTML]{D5CFCD}6.68}} & \multirow{-2}{*}{5.85} \\ \cline{2-11} 
\multicolumn{1}{c|}{} & \multicolumn{1}{c|}{} & \multicolumn{1}{c|}{} & \multicolumn{1}{c|}{\cellcolor[HTML]{D5CFCD}} & \multicolumn{1}{c|}{} & \multicolumn{1}{c|}{} & \multicolumn{1}{c|}{\cellcolor[HTML]{D5CFCD}} & \multicolumn{1}{c|}{} & \multicolumn{1}{c|}{} & \multicolumn{1}{c|}{\cellcolor[HTML]{D5CFCD}} &  \\
\multicolumn{1}{c|}{} & \multicolumn{1}{c|}{\multirow{-2}{*}{AdvIBP\tnote{2}}} & \multicolumn{1}{c|}{\multirow{-2}{*}{4.15}} & \multicolumn{1}{c|}{\multirow{-2}{*}{\cellcolor[HTML]{D5CFCD}10.80}} & \multicolumn{1}{c|}{\multirow{-2}{*}{6.83}} & \multicolumn{1}{c|}{\multirow{-2}{*}{2.33}} & \multicolumn{1}{c|}{\multirow{-2}{*}{\cellcolor[HTML]{D5CFCD}8.73}} & \multicolumn{1}{c|}{\multirow{-2}{*}{4.35}} & \multicolumn{1}{c|}{\multirow{-2}{*}{1.58}} & \multicolumn{1}{c|}{\multirow{-2}{*}{\cellcolor[HTML]{D5CFCD}8.23}} & \multirow{-2}{*}{3.17} \\ \cline{2-11} 
\multicolumn{1}{c|}{} & \multicolumn{1}{c|}{} & \multicolumn{1}{c|}{} & \multicolumn{1}{c|}{\cellcolor[HTML]{D5CFCD}} & \multicolumn{1}{c|}{} & \multicolumn{1}{c|}{} & \multicolumn{1}{c|}{\cellcolor[HTML]{D5CFCD}} & \multicolumn{1}{c|}{} & \multicolumn{1}{c|}{} & \multicolumn{1}{c|}{\cellcolor[HTML]{D5CFCD}} &  \\
\multicolumn{1}{c|}{\multirow{-6}{*}{$\epsilon=0.3$}} & \multicolumn{1}{c|}{\multirow{-2}{*}{AdvCROWN-IBP}} & \multicolumn{1}{c|}{\multirow{-2}{*}{3.22}} & \multicolumn{1}{c|}{\multirow{-2}{*}{\cellcolor[HTML]{D5CFCD}\textbf{9.03}}} & \multicolumn{1}{c|}{\multirow{-2}{*}{5.42}} & \multicolumn{1}{c|}{\multirow{-2}{*}{2.45}} & \multicolumn{1}{c|}{\multirow{-2}{*}{\cellcolor[HTML]{D5CFCD}\textbf{8.31}}} & \multicolumn{1}{c|}{\multirow{-2}{*}{3.81}} & \multicolumn{1}{c|}{\multirow{-2}{*}{1.90}} & \multicolumn{1}{c|}{\multirow{-2}{*}{\cellcolor[HTML]{D5CFCD}\textbf{6.60}}} & \multirow{-2}{*}{2.87} \\ \hline
\multicolumn{1}{c|}{} & \multicolumn{1}{c|}{} & \multicolumn{1}{c|}{} & \multicolumn{1}{c|}{\cellcolor[HTML]{D5CFCD}} & \multicolumn{1}{c|}{} & \multicolumn{1}{c|}{} & \multicolumn{1}{c|}{\cellcolor[HTML]{D5CFCD}} & \multicolumn{1}{c|}{} & \multicolumn{1}{c|}{} & \multicolumn{1}{c|}{\cellcolor[HTML]{D5CFCD}} &  \\
\multicolumn{1}{c|}{} & \multicolumn{1}{c|}{\multirow{-2}{*}{CROWN-IBP}} & \multicolumn{1}{c|}{\multirow{-2}{*}{3.78}} & \multicolumn{1}{c|}{\multirow{-2}{*}{\cellcolor[HTML]{D5CFCD}15.21}} & \multicolumn{1}{c|}{\multirow{-2}{*}{13.34}} & \multicolumn{1}{c|}{\multirow{-2}{*}{3.16}} & \multicolumn{1}{c|}{\multirow{-2}{*}{\cellcolor[HTML]{D5CFCD}14.19}} & \multicolumn{1}{c|}{\multirow{-2}{*}{11.31}} & \multicolumn{1}{c|}{\multirow{-2}{*}{1.62}} & \multicolumn{1}{c|}{\multirow{-2}{*}{\cellcolor[HTML]{D5CFCD}12.46}} & \multirow{-2}{*}{9.47} \\ \cline{2-11} 
\multicolumn{1}{c|}{} & \multicolumn{1}{c|}{} & \multicolumn{1}{c|}{} & \multicolumn{1}{c|}{\cellcolor[HTML]{D5CFCD}} & \multicolumn{1}{c|}{} & \multicolumn{1}{c|}{} & \multicolumn{1}{c|}{\cellcolor[HTML]{D5CFCD}} & \multicolumn{1}{c|}{} & \multicolumn{1}{c|}{} & \multicolumn{1}{c|}{\cellcolor[HTML]{D5CFCD}} &  \\
\multicolumn{1}{c|}{} & \multicolumn{1}{c|}{\multirow{-2}{*}{AdvIBP}} & \multicolumn{1}{c|}{\multirow{-2}{*}{4.15}} & \multicolumn{1}{c|}{\multirow{-2}{*}{\cellcolor[HTML]{D5CFCD}17.57}} & \multicolumn{1}{c|}{\multirow{-2}{*}{8.48}} & \multicolumn{1}{c|}{\multirow{-2}{*}{2.72}} & \multicolumn{1}{c|}{\multirow{-2}{*}{\cellcolor[HTML]{D5CFCD}16.18}} & \multicolumn{1}{c|}{\multirow{-2}{*}{5.58}} & \multicolumn{1}{c|}{\multirow{-2}{*}{1.88}} & \multicolumn{1}{c|}{\multirow{-2}{*}{\cellcolor[HTML]{D5CFCD}16.57}} & \multirow{-2}{*}{3.23} \\ \cline{2-11} 
\multicolumn{1}{c|}{} & \multicolumn{1}{c|}{} & \multicolumn{1}{c|}{} & \multicolumn{1}{c|}{\cellcolor[HTML]{D5CFCD}} & \multicolumn{1}{c|}{} & \multicolumn{1}{c|}{} & \multicolumn{1}{c|}{\cellcolor[HTML]{D5CFCD}} & \multicolumn{1}{c|}{} & \multicolumn{1}{c|}{} & \multicolumn{1}{c|}{\cellcolor[HTML]{D5CFCD}} &  \\
\multicolumn{1}{c|}{\multirow{-6}{*}{$\epsilon=0.4$}} & \multicolumn{1}{c|}{\multirow{-2}{*}{AdvCROWN-IBP}} & \multicolumn{1}{c|}{\multirow{-2}{*}{3.22}} & \multicolumn{1}{c|}{\multirow{-2}{*}{\cellcolor[HTML]{D5CFCD}\textbf{14.42}}} & \multicolumn{1}{c|}{\multirow{-2}{*}{6.69}} & \multicolumn{1}{c|}{\multirow{-2}{*}{2.98}} & \multicolumn{1}{c|}{\multirow{-2}{*}{\cellcolor[HTML]{D5CFCD}\textbf{13.88}}} & \multicolumn{1}{c|}{\multirow{-2}{*}{6.38}} & \multicolumn{1}{c|}{\multirow{-2}{*}{1.90}} & \multicolumn{1}{c|}{\multirow{-2}{*}{\cellcolor[HTML]{D5CFCD}\textbf{12.30}}} & \multirow{-2}{*}{3.46} \\ \bottomrule
\end{tabular}
\begin{tablenotes}
    \item[1] To further probe the \textit{true robustness} of the trained models, we verify the robustness of the \toolname trained models with a more precise method, RefineZono. The results are shown in Table~\ref{tab:post_robustness} in the Appendix.
    \item[2] We have also tested three model structures similar to DM-Small, DM-Medium and DM-Large. Results are reported in Table~\ref{tab:full_results} in the Appendix. For these models, \toolname already outperforms CROWN-IBP in all settings.
\end{tablenotes}
\end{threeparttable}
\end{adjustbox}
\label{tab:robustness}
\end{table*}


%% file: tex/conclusion.tex
We propose a new certified adversarial training framework that bridges the gap between adversarial training and provable robustness from a joint training perspective.
We formulate the joint training as a two-objective optimization problem,
which facilitates the balance between \textit{adversarial loss} and \textit{abstract loss}.
We show that our joint training framework outperforms prior certified adversarial training methods in both standard and verified errors, and achieves state-of-the-art verified test errors for $l_\infty$ robustness.

%% file: tex/appendix.tex
\setcounter{table}{0}
\setcounter{figure}{0}
\renewcommand{\thetable}{\Alph{table}}
\renewcommand{\thefigure}{\Alph{figure}}

\subsection{Additional Experiment Results}
Table~\ref{tab:full_results} presents the verified and standard errors for three model architectures in~\ref{tab:new_models} and perturbation settings used in~\citep{zhang2019towards}. On these three new models, \toolname outperform CROWN-IBP on the verified and standard errors in all settings.

\begin{table}[htp]
\centering
\caption{Model structures used in the experiments. "Conv $k$ $w \times h + s$" represents a 2D convolutional layer with $k$ filters of size $w \times h$ using a stride of $s$ in both dimensions. "FC $n$" represents a fully connected layer with $n$ outputs. The last fully connected layer is omitted. All networks use ReLU activation functions.}
\begin{adjustbox}{width=\columnwidth,center}
\begin{tabular}{@{}c|c|c@{}}
    \toprule
    CNN-Small                 & CNN-Medium                & CNN-Large                  \\ \midrule
    Conv 16 $3 \times 3 + 2$ & Conv 32 $3 \times 3 + 2$ & Conv 64 $3 \times 3 + 2$  \\
    Conv 32 $3 \times 3 + 1$ & Conv 64 $3 \times 3 + 1$ & Conv 64 $3 \times 3 + 1$  \\
    FC 512                   & Conv 128 $3 \times 3 + 1$ & Conv 128 $3 \times 3 + 1$ \\
                             & Conv 256 $3 \times 3 + 1$ & Conv 256 $3 \times 3 + 1$ \\
                             & FC 512                   & Conv 256 $3 \times 3 + 1$ \\
                             & FC 512                   & FC 512                    \\ \bottomrule
\end{tabular}
\end{adjustbox}
\label{tab:new_models}
\end{table}

\subsection{Post-training verification.}
A more precise verification usually comes at the cost of heavy computation (e.g. days vs seconds) and intensive memory usage compared with IBP, and is thus more suitable post-training.
In many cases, we observe that the PGD errors of the models trained by \toolname are smaller than the PGD errors of those trained by CROWN-IBP.
To understand and evaluate the ground-truth robustness, we apply RefineZono to verify the \toolname trained network.
Table~\ref{tab:post_robustness} shows the comparison on MNIST. We randomly sample 1000 images from MNIST test dataset which has $10000$ images in total.
RefineZono provides a tighter estimation of robust loss and smaller verified errors compared with results obtained from IBP. More interestingly, in more than half of the cases, \toolname obtains a lower verified error than the corresponding PGD error in CROWN-IBP. 
Since PGD error provides a lower bound on the robustness of the network, this means that those networks (results in bold) trained using \toolname are strictly more robust than those trained using CROWN-IBP with the same architecture and perturbation settings.

\subsection{The Computation of FOSC Value}
We use Eq.(4) in \citet{wang2019convergence} to compute the FOSC value for $\vx_\text{adv}$ generated by FGSM+random initialization. Specifically, for an allowable perturbation input set $\sS(\vx, \epsilon)$, the FOSC value of $\vx_\text{adv} \in \sS(\vx, \epsilon)$ is computed as follows.
\begin{equation*}
    c(\vx_\text{adv}) = \epsilon \Vert \nabla_\vx \gL(f_\theta(\vx_\text{adv}), y) \Vert_1 - \langle \vx_\text{adv} - x, \nabla_\vx \gL(f_\theta(\vx_\text{adv}), y) \rangle
\end{equation*}
For a mini-batch $\gB$, we compute the average of the FOSC values, $c(\vx){=}\sum_{i{=}0}^{|\gB| {-} 1} c(\vx_{\text{adv}, i})$, and compare it with $c_t$ to determine the prioritization direction (line 12 in \Algref{algorithm:two_obj_training}).

\subsection{Training Time}

In terms of \textbf{runtime} compared to the baseline method in Figure~\ref{fig:baseline}, the training times of the baseline method for the three models were 5942, 38290, and 108683 seconds. On the same models and hardware, \toolname took 6683, 43024, and 128747 seconds respectively.
This shows that the automatic coefficient computation in \toolname does not significantly increase training time.
Furthermore, since \toolname primarily combines FGSM-AT and IBP training, the efficiency of \toolname is comparable to those of FGSM-AT and IBP.

In Table~\ref{tab:runtime}, we present the training time of \toolname for the models used in our experiments. The hyperparameters are chosen as stated in~\ref{sec:hyperparamters}. We perform layer-wise training for each model and report the total training time here. All experiments are measured on a single GeForce GTX 1080 Ti GPU with 11 GB RAM except \textbf{DM-Large} models where we used 4  GeForce GTX 1080 Ti GPUs to speed up training.

\begin{table}[hp]
\centering
\caption{Average training time for different model architectures on each dataset.}
\begin{adjustbox}{width=.99\columnwidth,center}
\begin{tabular}{@{}cclclcl@{}}
    \toprule
    \multirow{2}{*}{Dataset} & \multicolumn{2}{c}{\multirow{2}{*}{DM-Small model (s)}} & \multicolumn{2}{c}{\multirow{2}{*}{DM-Model model (s)}} & \multicolumn{2}{c}{\multirow{2}{*}{DM-Large model (s)}} \\
                             & \multicolumn{2}{c}{}                                & \multicolumn{2}{c}{}                                & \multicolumn{2}{c}{}                                \\ \midrule
    MNIST                    & \multicolumn{2}{c}{6683}                            & \multicolumn{2}{c}{43024}                           & \multicolumn{2}{c}{128747}                          \\
    CIFAR-10                 & \multicolumn{2}{c}{63570}                           & \multicolumn{2}{c}{292863}                          & \multicolumn{2}{c}{527667}                          \\ \bottomrule
\end{tabular}
\end{adjustbox}
\label{tab:runtime}
\end{table}

\begin{table*}[htpb]
\caption{The standard and verified errors for trained models on MNIST.}
\begin{adjustbox}{width=\textwidth,center}
\begin{threeparttable}
\begin{tabular}{cccccccc}
\toprule
 &  & \multicolumn{2}{c}{} & \multicolumn{2}{c}{} & \multicolumn{2}{c}{} \\
 &  & \multicolumn{2}{c}{\multirow{-2}{*}{CNN-Small model's err. (\%)}} & \multicolumn{2}{c}{\multirow{-2}{*}{CNN-Medium model's err. (\%)}} & \multicolumn{2}{c}{\multirow{-2}{*}{CNN-Large model's err. (\%)}} \\ \cline{3-8} 
 &  & \multicolumn{1}{c|}{} & \multicolumn{1}{c|}{\cellcolor[HTML]{D5CFCD}} & \multicolumn{1}{c|}{} & \multicolumn{1}{c|}{\cellcolor[HTML]{D5CFCD}} & \multicolumn{1}{c|}{} & \cellcolor[HTML]{D5CFCD} \\
\multirow{-4}{*}{$\epsilon$ ($\evl_\infty$ norm)} & \multirow{-4}{*}{Training Method} & \multicolumn{1}{c|}{\multirow{-2}{*}{Standard}} & \multicolumn{1}{c|}{\multirow{-2}{*}{\cellcolor[HTML]{D5CFCD}Verified}} & \multicolumn{1}{c|}{\multirow{-2}{*}{Standard}} & \multicolumn{1}{c|}{\multirow{-2}{*}{\cellcolor[HTML]{D5CFCD}Verified}} & \multicolumn{1}{c|}{\multirow{-2}{*}{Standard}} & \multirow{-2}{*}{\cellcolor[HTML]{D5CFCD}Verified} \\ \hline
\multicolumn{1}{c|}{} & \multicolumn{1}{c|}{} & \multicolumn{1}{c|}{} & \multicolumn{1}{c|}{\cellcolor[HTML]{D5CFCD}} & \multicolumn{1}{c|}{} & \multicolumn{1}{c|}{\cellcolor[HTML]{D5CFCD}} & \multicolumn{1}{c|}{} & \cellcolor[HTML]{D5CFCD} \\
\multicolumn{1}{c|}{} & \multicolumn{1}{c|}{\multirow{-2}{*}{CROWN-IBP}} & \multicolumn{1}{c|}{\multirow{-2}{*}{1.44}} & \multicolumn{1}{c|}{\multirow{-2}{*}{\cellcolor[HTML]{D5CFCD}2.85}} & \multicolumn{1}{c|}{\multirow{-2}{*}{1.19}} & \multicolumn{1}{c|}{\multirow{-2}{*}{\cellcolor[HTML]{D5CFCD}2.65}} & \multicolumn{1}{c|}{\multirow{-2}{*}{1.07}} & \multirow{-2}{*}{\cellcolor[HTML]{D5CFCD}2.56} \\ \cline{2-8} 
\multicolumn{1}{c|}{} & \multicolumn{1}{c|}{} & \multicolumn{1}{c|}{} & \multicolumn{1}{c|}{\cellcolor[HTML]{D5CFCD}} & \multicolumn{1}{c|}{} & \multicolumn{1}{c|}{\cellcolor[HTML]{D5CFCD}} & \multicolumn{1}{c|}{} & \cellcolor[HTML]{D5CFCD} \\
\multicolumn{1}{c|}{\multirow{-4}{*}{$\epsilon=0.1$}} & \multicolumn{1}{c|}{\multirow{-2}{*}{AdvIBP}} & \multicolumn{1}{c|}{\multirow{-2}{*}{1.09}} & \multicolumn{1}{c|}{\multirow{-2}{*}{\cellcolor[HTML]{D5CFCD}\textbf{2.63}}} & \multicolumn{1}{c|}{\multirow{-2}{*}{1.12}} & \multicolumn{1}{c|}{\multirow{-2}{*}{\cellcolor[HTML]{D5CFCD}\textbf{2.60}}} & \multicolumn{1}{c|}{\multirow{-2}{*}{1.01}} & \multirow{-2}{*}{\cellcolor[HTML]{D5CFCD}\textbf{2.50}} \\ \hline
\multicolumn{1}{c|}{} & \multicolumn{1}{c|}{} & \multicolumn{1}{c|}{} & \multicolumn{1}{c|}{\cellcolor[HTML]{D5CFCD}} & \multicolumn{1}{c|}{} & \multicolumn{1}{c|}{\cellcolor[HTML]{D5CFCD}} & \multicolumn{1}{c|}{} & \cellcolor[HTML]{D5CFCD} \\
\multicolumn{1}{c|}{} & \multicolumn{1}{c|}{\multirow{-2}{*}{CROWN-IBP}} & \multicolumn{1}{c|}{\multirow{-2}{*}{2.81}} & \multicolumn{1}{c|}{\multirow{-2}{*}{\cellcolor[HTML]{D5CFCD}5.79}} & \multicolumn{1}{c|}{\multirow{-2}{*}{2.57}} & \multicolumn{1}{c|}{\multirow{-2}{*}{\cellcolor[HTML]{D5CFCD}4.93}} & \multicolumn{1}{c|}{\multirow{-2}{*}{2.34}} & \multirow{-2}{*}{\cellcolor[HTML]{D5CFCD}4.71} \\ \cline{2-8} 
\multicolumn{1}{c|}{} & \multicolumn{1}{c|}{} & \multicolumn{1}{c|}{} & \multicolumn{1}{c|}{\cellcolor[HTML]{D5CFCD}} & \multicolumn{1}{c|}{} & \multicolumn{1}{c|}{\cellcolor[HTML]{D5CFCD}} & \multicolumn{1}{c|}{} & \cellcolor[HTML]{D5CFCD} \\
\multicolumn{1}{c|}{\multirow{-4}{*}{$\epsilon=0.2$}} & \multicolumn{1}{c|}{\multirow{-2}{*}{AdvIBP}} & \multicolumn{1}{c|}{\multirow{-2}{*}{2.38}} & \multicolumn{1}{c|}{\multirow{-2}{*}{\cellcolor[HTML]{D5CFCD}\textbf{5.22}}} & \multicolumn{1}{c|}{\multirow{-2}{*}{2.33}} & \multicolumn{1}{c|}{\multirow{-2}{*}{\cellcolor[HTML]{D5CFCD}\textbf{4.66}}} & \multicolumn{1}{c|}{\multirow{-2}{*}{2.04}} & \multirow{-2}{*}{\cellcolor[HTML]{D5CFCD}\textbf{4.37}} \\ \hline
\multicolumn{1}{c|}{} & \multicolumn{1}{c|}{} & \multicolumn{1}{c|}{} & \multicolumn{1}{c|}{\cellcolor[HTML]{D5CFCD}} & \multicolumn{1}{c|}{} & \multicolumn{1}{c|}{\cellcolor[HTML]{D5CFCD}} & \multicolumn{1}{c|}{} & \cellcolor[HTML]{D5CFCD} \\
\multicolumn{1}{c|}{} & \multicolumn{1}{c|}{\multirow{-2}{*}{CROWN-IBP}} & \multicolumn{1}{c|}{\multirow{-2}{*}{2.81}} & \multicolumn{1}{c|}{\multirow{-2}{*}{\cellcolor[HTML]{D5CFCD}8.55}} & \multicolumn{1}{c|}{\multirow{-2}{*}{2.57}} & \multicolumn{1}{c|}{\multirow{-2}{*}{\cellcolor[HTML]{D5CFCD}7.78}} & \multicolumn{1}{c|}{\multirow{-2}{*}{2.34}} & \multirow{-2}{*}{\cellcolor[HTML]{D5CFCD}7.22} \\ \cline{2-8} 
\multicolumn{1}{c|}{} & \multicolumn{1}{c|}{} & \multicolumn{1}{c|}{} & \multicolumn{1}{c|}{\cellcolor[HTML]{D5CFCD}} & \multicolumn{1}{c|}{} & \multicolumn{1}{c|}{\cellcolor[HTML]{D5CFCD}} & \multicolumn{1}{c|}{} & \cellcolor[HTML]{D5CFCD} \\
\multicolumn{1}{c|}{\multirow{-4}{*}{$\epsilon=0.3$}} & \multicolumn{1}{c|}{\multirow{-2}{*}{AdvIBP}} & \multicolumn{1}{c|}{\multirow{-2}{*}{2.38}} & \multicolumn{1}{c|}{\multirow{-2}{*}{\cellcolor[HTML]{D5CFCD}\textbf{8.52}}} & \multicolumn{1}{c|}{\multirow{-2}{*}{2.33}} & \multicolumn{1}{c|}{\multirow{-2}{*}{\cellcolor[HTML]{D5CFCD}\textbf{7.58}}} & \multicolumn{1}{c|}{\multirow{-2}{*}{2.04}} & \multirow{-2}{*}{\cellcolor[HTML]{D5CFCD}\textbf{6.87}} \\ \hline
\multicolumn{1}{c|}{} & \multicolumn{1}{c|}{} & \multicolumn{1}{c|}{} & \multicolumn{1}{c|}{\cellcolor[HTML]{D5CFCD}} & \multicolumn{1}{c|}{} & \multicolumn{1}{c|}{\cellcolor[HTML]{D5CFCD}} & \multicolumn{1}{c|}{} & \cellcolor[HTML]{D5CFCD} \\
\multicolumn{1}{c|}{} & \multicolumn{1}{c|}{\multirow{-2}{*}{CROWN-IBP}} & \multicolumn{1}{c|}{\multirow{-2}{*}{2.81}} & \multicolumn{1}{c|}{\multirow{-2}{*}{\cellcolor[HTML]{D5CFCD}13.74}} & \multicolumn{1}{c|}{\multirow{-2}{*}{2.57}} & \multicolumn{1}{c|}{\multirow{-2}{*}{\cellcolor[HTML]{D5CFCD}13.53}} & \multicolumn{1}{c|}{\multirow{-2}{*}{2.34}} & \multirow{-2}{*}{\cellcolor[HTML]{D5CFCD}12.00} \\ \cline{2-8} 
\multicolumn{1}{c|}{} & \multicolumn{1}{c|}{} & \multicolumn{1}{c|}{} & \multicolumn{1}{c|}{\cellcolor[HTML]{D5CFCD}} & \multicolumn{1}{c|}{} & \multicolumn{1}{c|}{\cellcolor[HTML]{D5CFCD}} & \multicolumn{1}{c|}{} & \cellcolor[HTML]{D5CFCD} \\
\multicolumn{1}{c|}{\multirow{-4}{*}{$\epsilon=0.4$}} & \multicolumn{1}{c|}{\multirow{-2}{*}{AdvIBP}} & \multicolumn{1}{c|}{\multirow{-2}{*}{2.38}} & \multicolumn{1}{c|}{\multirow{-2}{*}{\cellcolor[HTML]{D5CFCD}\textbf{13.24}}} & \multicolumn{1}{c|}{\multirow{-2}{*}{2.33}} & \multicolumn{1}{c|}{\multirow{-2}{*}{\cellcolor[HTML]{D5CFCD}\textbf{13.36}}} & \multicolumn{1}{c|}{\multirow{-2}{*}{2.04}} & \multirow{-2}{*}{\cellcolor[HTML]{D5CFCD}\textbf{11.98}} \\ \bottomrule
\end{tabular}
\end{threeparttable}
\end{adjustbox}
\label{tab:full_results}
\end{table*}

\begin{table*}[htpb]
\caption[Caption for LOF]{Evaluation on the MNIST dataset between models trained by \toolname and those by CROWN-IBP. We verify the networks with RefineZono~\citep{singh2019boosting} for 1000 images randomly sampled from the test dataset. The IBP verified errors are in parentheses. The CROWN-IBP result\footnotemark is from Table C. in~\citep{zhang2019towards}.}
\begin{adjustbox}{width=0.9\textwidth,center}
\begin{threeparttable}
\begin{tabular}{cccccccc}
\toprule
 &  & \multicolumn{2}{c}{} & \multicolumn{2}{c}{} & \multicolumn{2}{c}{} \\
 &  & \multicolumn{2}{c}{\multirow{-2}{*}{DM-Small model's err. (\%)}} & \multicolumn{2}{c}{\multirow{-2}{*}{DM-Medium model's err. (\%)}} & \multicolumn{2}{c}{\multirow{-2}{*}{DM-Large model's err. (\%)}} \\ \cline{3-8} 
 &  & \multicolumn{1}{c|}{} & \multicolumn{1}{c|}{} & \multicolumn{1}{c|}{} & \multicolumn{1}{c|}{} & \multicolumn{1}{c|}{} &  \\
\multirow{-4}{*}{$\epsilon$ ($\evl_\infty$ norm)\tnote{2}} & \multirow{-4}{*}{Training Method} & \multicolumn{1}{c|}{\multirow{-2}{*}{Verified}} & \multicolumn{1}{c|}{\multirow{-2}{*}{PGD}} & \multicolumn{1}{c|}{\multirow{-2}{*}{Verified}} & \multicolumn{1}{c|}{\multirow{-2}{*}{PGD}} & \multicolumn{1}{c|}{\multirow{-2}{*}{Verified}} & \multirow{-2}{*}{PGD} \\ \hline
\multicolumn{1}{c|}{} & \multicolumn{1}{c|}{} & \multicolumn{1}{c|}{\cellcolor[HTML]{D5CFCD}} & \multicolumn{1}{c|}{} & \multicolumn{1}{c|}{\cellcolor[HTML]{D5CFCD}} & \multicolumn{1}{c|}{} & \multicolumn{1}{c|}{\cellcolor[HTML]{D5CFCD}} &  \\
\multicolumn{1}{c|}{} & \multicolumn{1}{c|}{\multirow{-2}{*}{CROWN-IBP}} & \multicolumn{1}{c|}{\multirow{-2}{*}{\cellcolor[HTML]{D5CFCD}3.44}} & \multicolumn{1}{c|}{\multirow{-2}{*}{3.09}} & \multicolumn{1}{c|}{\multirow{-2}{*}{\cellcolor[HTML]{D5CFCD}3.26}} & \multicolumn{1}{c|}{\multirow{-2}{*}{2.82}} & \multicolumn{1}{c|}{\multirow{-2}{*}{\cellcolor[HTML]{D5CFCD} 2.10 (2.24)}} & \multirow{-2}{*}{1.81} \\ \cline{2-8} 
\multicolumn{1}{c|}{} & \multicolumn{1}{c|}{} & \multicolumn{1}{c|}{\cellcolor[HTML]{D5CFCD}} & \multicolumn{1}{c|}{} & \multicolumn{1}{c|}{\cellcolor[HTML]{D5CFCD}} & \multicolumn{1}{c|}{} & \multicolumn{1}{c|}{\cellcolor[HTML]{D5CFCD}} &  \\
\multicolumn{1}{c|}{\multirow{-4}{*}{$\epsilon=0.1$}} & \multicolumn{1}{c|}{\multirow{-2}{*}{\toolname}} & \multicolumn{1}{c|}{\multirow{-2}{*}{\cellcolor[HTML]{D5CFCD}\textbf{2.90 }(3.69)}} & \multicolumn{1}{c|}{\multirow{-2}{*}{2.70}} & \multicolumn{1}{c|}{\multirow{-2}{*}{\cellcolor[HTML]{D5CFCD}\textbf{2.80} (3.24)}} & \multicolumn{1}{c|}{\multirow{-2}{*}{2.26}} & \multicolumn{1}{c|}{\multirow{-2}{*}{\cellcolor[HTML]{D5CFCD}\textbf{1.90} (2.28)}} & \multirow{-2}{*}{1.53} \\ \hline
\multicolumn{1}{c|}{} & \multicolumn{1}{c|}{} & \multicolumn{1}{c|}{\cellcolor[HTML]{D5CFCD}} & \multicolumn{1}{c|}{} & \multicolumn{1}{c|}{\cellcolor[HTML]{D5CFCD}} & \multicolumn{1}{c|}{} & \multicolumn{1}{c|}{\cellcolor[HTML]{D5CFCD}} &  \\
\multicolumn{1}{c|}{} & \multicolumn{1}{c|}{\multirow{-2}{*}{CROWN-IBP}} & \multicolumn{1}{c|}{\multirow{-2}{*}{\cellcolor[HTML]{D5CFCD}6.11}} & \multicolumn{1}{c|}{\multirow{-2}{*}{5.74}} & \multicolumn{1}{c|}{\multirow{-2}{*}{\cellcolor[HTML]{D5CFCD}5.35}} & \multicolumn{1}{c|}{\multirow{-2}{*}{4.90}} & \multicolumn{1}{c|}{\multirow{-2}{*}{\cellcolor[HTML]{D5CFCD}4.00 (3.87)}} & \multirow{-2}{*}{3.81} \\ \cline{2-8} 
\multicolumn{1}{c|}{} & \multicolumn{1}{c|}{} & \multicolumn{1}{c|}{\cellcolor[HTML]{D5CFCD}} & \multicolumn{1}{c|}{} & \multicolumn{1}{c|}{\cellcolor[HTML]{D5CFCD}} & \multicolumn{1}{c|}{} & \multicolumn{1}{c|}{\cellcolor[HTML]{D5CFCD}} &  \\
\multicolumn{1}{c|}{\multirow{-4}{*}{$\epsilon=0.2$}} & \multicolumn{1}{c|}{\multirow{-2}{*}{\toolname}} & \multicolumn{1}{c|}{\multirow{-2}{*}{\cellcolor[HTML]{D5CFCD}6.80 (7.68)}} & \multicolumn{1}{c|}{\multirow{-2}{*}{5.81}} & \multicolumn{1}{c|}{\multirow{-2}{*}{\cellcolor[HTML]{D5CFCD}\textbf{4.50 }(5.37)}} & \multicolumn{1}{c|}{\multirow{-2}{*}{3.54}} & \multicolumn{1}{c|}{\multirow{-2}{*}{\cellcolor[HTML]{D5CFCD}\textbf{3.80 }(4.70)}} & \multirow{-2}{*}{2.59} \\ \hline
\multicolumn{1}{c|}{} & \multicolumn{1}{c|}{} & \multicolumn{1}{c|}{\cellcolor[HTML]{D5CFCD}} & \multicolumn{1}{c|}{} & \multicolumn{1}{c|}{\cellcolor[HTML]{D5CFCD}} & \multicolumn{1}{c|}{} & \multicolumn{1}{c|}{\cellcolor[HTML]{D5CFCD}} &  \\
\multicolumn{1}{c|}{} & \multicolumn{1}{c|}{\multirow{-2}{*}{CROWN-IBP}} & \multicolumn{1}{c|}{\multirow{-2}{*}{\cellcolor[HTML]{D5CFCD}9.40}} & \multicolumn{1}{c|}{\multirow{-2}{*}{8.50}} & \multicolumn{1}{c|}{\multirow{-2}{*}{\cellcolor[HTML]{D5CFCD}8.54}} & \multicolumn{1}{c|}{\multirow{-2}{*}{7.74}} & \multicolumn{1}{c|}{\multirow{-2}{*}{\cellcolor[HTML]{D5CFCD} 6.20\tnote{3} (6.68)}} & \multirow{-2}{*}{5.85} \\ \cline{2-8} 
\multicolumn{1}{c|}{} & \multicolumn{1}{c|}{} & \multicolumn{1}{c|}{\cellcolor[HTML]{D5CFCD}} & \multicolumn{1}{c|}{} & \multicolumn{1}{c|}{\cellcolor[HTML]{D5CFCD}} & \multicolumn{1}{c|}{} & \multicolumn{1}{c|}{\cellcolor[HTML]{D5CFCD}} &  \\
\multicolumn{1}{c|}{\multirow{-4}{*}{$\epsilon=0.3$}} & \multicolumn{1}{c|}{\multirow{-2}{*}{\toolname}} & \multicolumn{1}{c|}{\multirow{-2}{*}{\cellcolor[HTML]{D5CFCD}9.30 (10.80)}} & \multicolumn{1}{c|}{\multirow{-2}{*}{6.83}} & \multicolumn{1}{c|}{\multirow{-2}{*}{\cellcolor[HTML]{D5CFCD}\textbf{6.80 }(8.73)}} & \multicolumn{1}{c|}{\multirow{-2}{*}{4.35}} & \multicolumn{1}{c|}{\multirow{-2}{*}{\cellcolor[HTML]{D5CFCD}\textbf{5.70}\tnote{3} (8.23)}} & \multirow{-2}{*}{3.17} \\ \bottomrule
\end{tabular}
\begin{tablenotes}
    \item[1] The models reported in~\citep{zhang2019learning} were unavailable except \textbf{DM-Large} at the time when this paper was written. As a result, we only report the verified errors obtained by RefineZono for the DM-Large models trained using CROWN-IBP. 
    \item[2] We does not present the result of RefineZono at $\epsilon{=}0.4$ due to the intensive GPU memory request. This configuration is not reported in related works~\citep{balunovic2020adversarial,mirman2018differentiable,singh2018fast} on similar-scale models.
    \item[3] The computation of the verified error under this setting uses DeepZ on a single TESLA V100 GPU with 16GB memory and costs 5 hours. RefineZono will take days to verify 1000 inputs.
\end{tablenotes}
\end{threeparttable}
\end{adjustbox}
\label{tab:post_robustness}
\end{table*}


\clearpage

\subsection{Models and Hyperparameters}
\label{sec:hyperparamters}
The models structures (\textbf{DM-Small}, \textbf{DM-Medium} and \textbf{DM-Large}) used in Table~\ref{tab:results} and \ref{tab:robustness} are listed in Table~\ref{tab:models}. These three model structures are the same as those in \citep{gowal2018effectiveness, zhang2019towards}. The large models are trained on 4-GPUs. For small and medium sized models, we train them on a single GPU.
The model structures used in Table~\ref{tab:all_models_results} are listed in Table~\ref{tab:all_models}. These models are all trained on a single GPU.
Training hyperparameters are detailed below:
\begin{itemize}
	\item For MNIST \toolname with $\epsilon_\text{train} = 0.2$ and $\epsilon_\text{train} = 0.4$ in Table~\ref{tab:robustness}, we set warm-up epochs as $T_\text{nat} = 10$ and $T_\text{adv} = 40$. We use Adam optimizer and set learning rate to $5 \times 10^{-4}$. After warm-up phase, we train $200$ epochs with a batch size of 256 and gradually ramp up $\epsilon$ from $0$ to $\epsilon_\text{train}$ in $R=50$ epochs with extra $10$ epochs training at $\epsilon_\text{train}$. We control the linear decay rate of $c_t$ with $T'$.
	We set $T' = 50$ starting with $c_\text{max} {=} 1e{-}4$ for $\epsilon_\text{train} {=} 0.4$ and $c_\text{max} {=} 1e{-}5$ for $\epsilon_\text{train} {=} 0.2$.
	We reduce the learning rate by $4\times$ at epoch $150$ and $200$. At the layer-wise training stage, we start with learning rate $1.25 \times 10^{-4}$ and train $250$ epochs each layer. We reduce the learning rate by $4\times$ at $50$, $100$ and $200$ epoch. The exponential decay rates for computing the weights are set to $\beta_1 = 0.9$ and $\beta_2 = 0.99$.
	\item For CIFAR-10 \toolname, we set $\epsilon_\text{train} = 1.1 \epsilon$ and train $3200$ epochs each layer with a batch size of 256. We set the first $800$ epochs as warm-up phase with $T_\text{nat} = 400$ and $T_\text{adv} = 410$. Then, we ramp up $\epsilon$ for $R = 1000$ epochs with extra $20$ epochs training at $\epsilon_\text{train}$.
	We set $T' = 280$ starting with $c_\text{max} {=} 1e{-}3$ for $\epsilon_\text{train} {=} \frac{8}{255}$ and $\epsilon {=} \frac{16}{255}$, and $c_\text{max} {=} 1e{-}5$ for $\epsilon {=} \frac{2}{255}$.
	Learning rate is reduced by $10\times$ at epoch $2200$ and $2700$ from $5 \times 10^{-4}$.
	At the layer-wise training stage, we start with learning rate $5 \times 10^{-5}$ and train $3200$ epochs each layer. We reduce the learning rate by $10\times$ every $1000$ epochs. The exponential decay rates for computing the weights are set to $\beta_1 = 0.9$ and $\beta_2 = 0.99$.
\end{itemize}

\begin{table}[htp]
\centering
\caption{Model structures used in the experiments. "Conv $k$ $w \times h + s$" represents a 2D convolutional layer with $k$ filters of size $w \times h$ using a stride of $s$ in both dimensions. "FC $n$" represents a fully connected layer with $n$ outputs. The last fully connected layer is omitted. All networks use ReLU activation functions.}
\begin{adjustbox}{width=\columnwidth,center}
\begin{tabular}{@{}c|c|c@{}}
    \toprule
    DM-Small                 & DM-Medium                & DM-Large                  \\ \midrule
    Conv 16 $4 \times 4 + 2$ & Conv 32 $3 \times 3 + 1$ & Conv 64 $3 \times 3 + 1$  \\
    Conv 32 $4 \times 4 + 1$ & Conv 32 $4 \times 4 + 2$ & Conv 64 $3 \times 3 + 1$  \\
    FC 100                   & Conv 64 $3 \times 3 + 1$ & Conv 128 $3 \times 3 + 2$ \\
                             & Conv 64 $4 \times 4 + 2$ & Conv 128 $3 \times 3 + 1$ \\
                             & FC 512                   & Conv 128 $3 \times 3 + 1$ \\
                             & FC 512                   & FC 512                    \\ \bottomrule
\end{tabular}
\end{adjustbox}
\label{tab:models}
\end{table}

\begin{table*}[htp]
\centering
\caption{10 Model structures used in Table~\ref{tab:all_models_results}. "Conv $k$ $w \times h + s$" represents a 2D convolutional layer with $k$ filters of size $w \times h$ using a stride of $s$ in both dimensions. "FC $n$" represents a fully connected layer with $n$ outputs. The last fully connected layer is omitted. All networks use ReLU activation functions.}
\begin{adjustbox}{width=\textwidth,center}
\begin{tabular}{c|c}
\toprule
Name & Model Strucutre \\ \hline
A & Conv 8 $3 \times 3 + 2$ , Conv 16 $3 \times 3 + 1$, FC 100 \\
B & Conv 16 $3 \times 3 + 2$ , Conv 32 $3 \times 3 + 1$, FC 100 \\
C & Conv 32 $3 \times 3 + 2$ , Conv 64 $3 \times 3 + 1$, FC 100 \\
D & Conv 8 $4 \times 4 + 2$ , Conv 16 $4 \times 4+ 1$, FC 512 \\
E & Conv 16 $4 \times 4 + 2$ , Conv 32 $4 \times 4 + 1$, FC 512 \\
F & Conv 32 $4 \times 4+ 2$ , Conv 64 $4 \times 4 + 1$, FC 512 \\
G & Conv 8 $3 \times 3 + 2$ , Conv 16 $3\times 3 + 1$, Conv 32 $3 \times 3 + 1$ , Conv 64 $3\times 3 + 1$, FC 512 \\
H (MNIST Only) & Conv 16 $3 \times 3 + 2$ , Conv 32 $3\times 3 + 1$, Conv 64 $3 \times 3 + 1$ , Conv 128 $3\times 3 + 1$, FC 512 \\
I (MNIST Only) & Conv 8 $3 \times 3 + 1$ , Conv 8 $4\times 4+ 2$, Conv 16 $3 \times 3 + 1$ , Conv 16 $4\times 4+ 2$, FC 512 \\
J (MNIST Only) & Conv 16 $3 \times 3 + 1$ , Conv 16 $4\times 4+ 2$, Conv 32 $3 \times 3 + 1$ , Conv 32 $4\times 4+ 2$, FC 512 \\ \bottomrule
\end{tabular}
\end{adjustbox}
\label{tab:all_models}
\end{table*}

\textbf{Hyperparameters $c_\text{max}$ and $\beta$.}
The hyperparameters unique to \toolname are the initial bound for adversarial strength (FOSC value) and the exponential moving average decay rate in moment estimates.

We choose the $c_\text{max}$ based on the adversarially trained networks. Let us assume that we fix the adversarial attack strategy to generate the perturbed inputs. For arbitrary networks without adversarial training, it is easy for a perturbed input to flip the output. Though the adversarial strength of the perturbed input is high (small FOSC value), the true robustness of the network is low. For adversarially trained networks, it becomes hard to find a perturbed input that can flip the output since these networks have strong empirical robustness.
The adversarial loss computed based on the generated perturbed inputs cannot approximate the robust loss tightly. As a result, a perturbed input can have low adversarial strength. The FOSC value of a perturbed input against an adversarially trained network provides an estimate of the largest FOSC value that an perturbed input can achieve under the specified perturbation against a robust model. At the beginning of training, this estimate allow us to train on weak adversarial examples to stabilize the training process.
\citet{wang2019convergence} observes that training on adversarial examples of higher adversarial strength at later stages leads to higher robustness. Thus, we set $c_\text{max}$ to the FOSC value based on the adversarially trained network initially. Then, after the warm-up phase, we gradually decrease $c_t$. Decreasing $c_t$ requires stronger and stronger adversarial examples to fulfill the condition $c(\vx_\text{adv}) \le c_t$ where the optimization prioritizes the minimization of adversarial loss.


Our choice of exponential average decay rate is inspired by the choices made in \citep{kingma2014adam}. The decay rate for the first order moment estimate is set to $\beta_1 = 0.9$. The decay rate for the 2-norm moment estimate is set to $\beta_2 = 0.99$. 
In the future, we plan to experiment with more choices for the decay rates to investigate how they influence the training process.

\subsection{Layer-wise Training}

\begin{figure}[b]
	\centering
	\subfloat[][MNIST, $\epsilon=0.3$]{
		\includegraphics[width=.47\columnwidth]{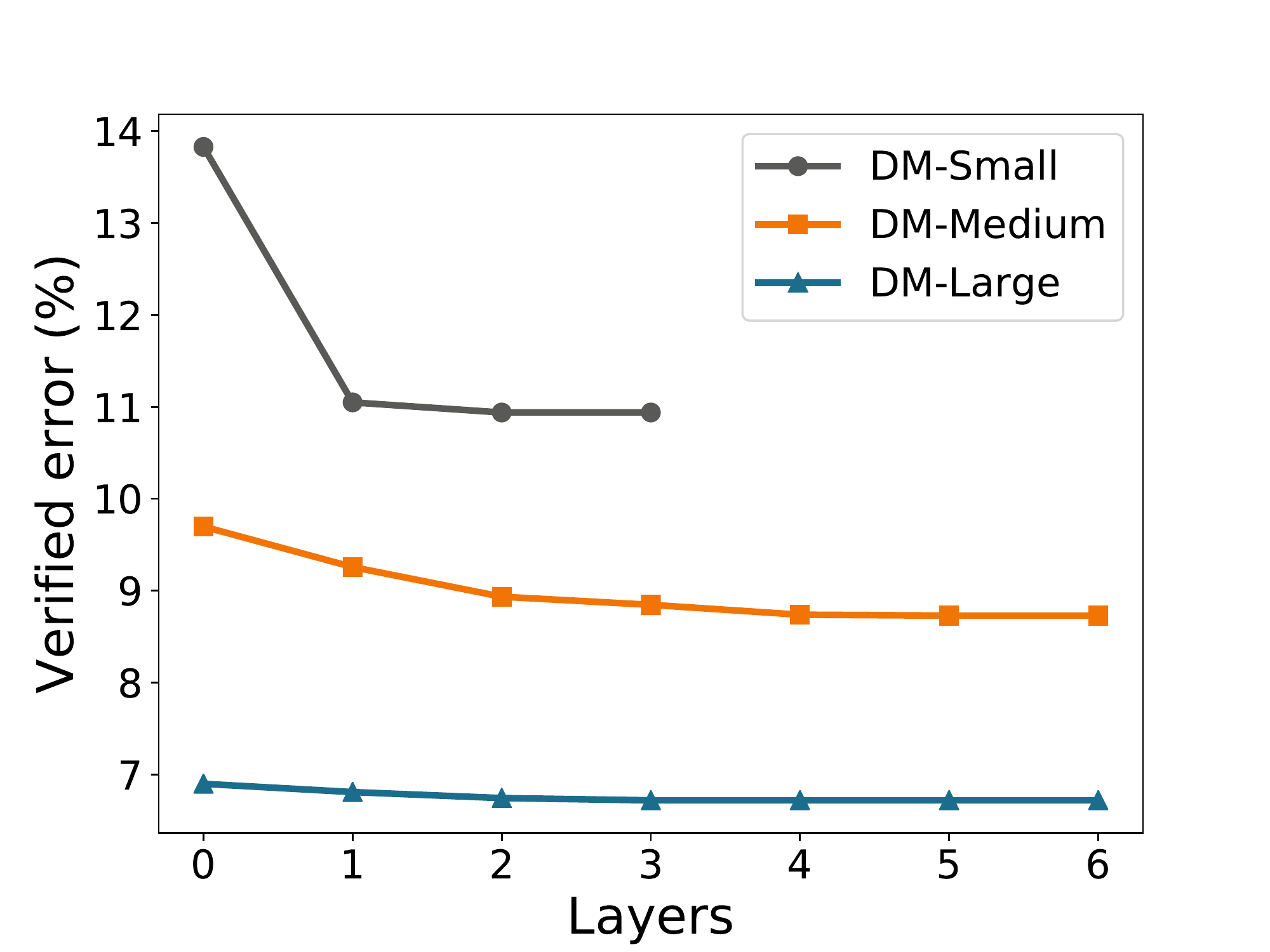}
		\label{fig:layerwise_mnist}
	}
	\hfill
	\subfloat[][CIFAR-10, $\epsilon=8/255$]{
		\includegraphics[width=.47\columnwidth]{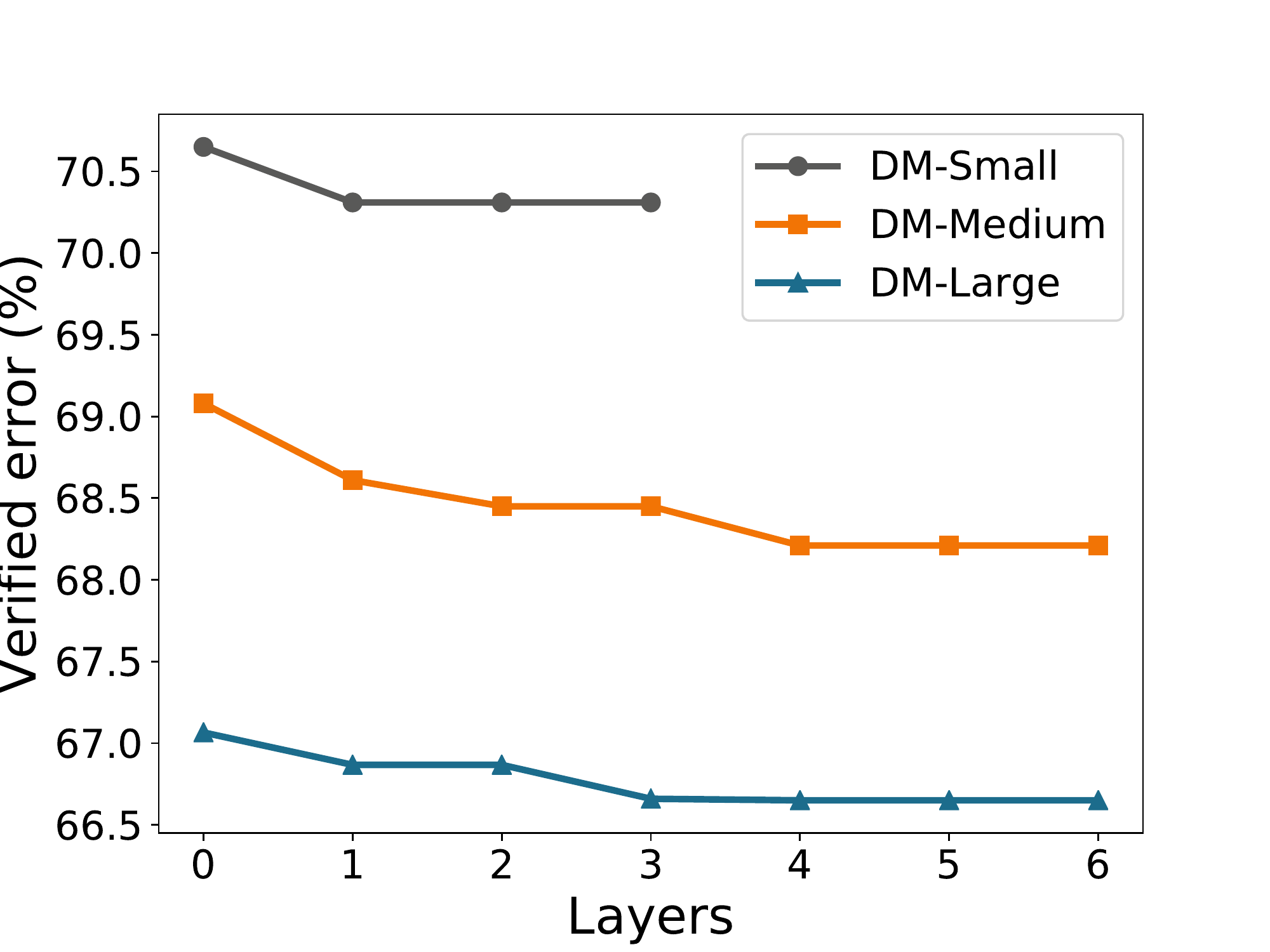}
		\label{fig:layerwise_cifar}
	}
	\caption{Effects of layer-wise training. The point at layer 0 represents standard adversarial training.}
	\label{fig:layerwise_effect}
\end{figure}

We investigate the role of latent adversarial examples plays in our training process. Recall that adversarial loss and latent adversarial loss are obtained from random initialized FGSM. 
We freeze the current layer where attacks apply and train the rest of the network layer by layer. 
In \Figref{fig:layerwise_effect}, we show the verified errors achieved when the parameters of the corresponding layer are frozen. 
Note that while layer-wise latent adversarial loss helps with improving certified robustness,
\toolname already produces competitive results in terms of verified error at layer 0 (i.e. adversarial examples in the inputs).
In addition, the layer-wise training strategy produces most of the improvement in the first two layers.
One major difference between \toolname and COLT~\citep{balunovic2020adversarial} is that we consider the tightness of over-approximation explicitly and model the trade-off between latent adversarial loss and abstract loss without requiring weight tuning. 
COLT~\citep{balunovic2020adversarial} does not explicitly consider abstract loss but use multiple regularization terms to control the over-approximation. 
Adding these regularization terms does not guarantee tighter over-approximation and the coefficient for each regularization term needs to be picked manually.

During the experiments, we also observed that strong latent adversarial examples are easy to be found if the over-approximation is loose. These latent adversarial examples are far from the boundary of the over-approximation. 
This indicates that minimizing the latent adversarial loss alone cannot guarantee better precision of over-approximation. 
This observation further motivates the need to explicitly consider the minimization of the abstract loss and our proposed joint training scheme.

\begin{strip}

\subsection{Proof of Theorem~\ref{theorem: same_dir} and \ref{theorem: diff_dir}}
\begin{assumption}
    The \textit{robust loss} and \textit{abstract loss} are locally strongly concave with respect to $\mu$ and $\overline{\mu}$ in allowable adversarial attack set $\sS(\vx, \epsilon)$. With the result of Lemma 1 in~\citep{wang2019convergence}, we have for any $\theta_1$ and $\theta_2$ the following holds
    \begin{equation*}
    \begin{array}{cc}
        \gL(\theta_1) {\le} \gL(\theta_2) {+} \langle \nabla \gL(\theta_2), \theta_1 {-} \theta_2 \rangle {+} \frac{L}{2} \Vert \theta_1 {-} \theta_2 \Vert_2^2, &
        \overline{\gL}(\theta_1) {\le} \overline{\gL}(\theta_2) {+} \langle \nabla \overline{\gL}(\theta_2), \theta_1 {-} \theta_2 \rangle {+} \frac{\overline{L}}{2} \Vert \theta_1 {-} \theta_2 \Vert_2^2
    \end{array}
    \end{equation*}
    where $L = (L_{\theta \vx} L_{\vx \theta} / \mu + L_{\theta \theta})$ and $\overline{L} = (\overline{L}_{\theta \vx} \overline{L}_{\vx \theta} / \overline{\mu} + \overline{L}_{\theta \theta})$.
    \label{assumption: concave}
\end{assumption}

The relation between (\ref{eq: minimax}) and distributional robust optimization~\citep{sinha2017certifying, lee2018minimax} supports the strongly concave assumption. The last assumption is common in stochastic gradient based optimization algorithms.

\begin{assumption}
    The variances of stochastic gradients $g(\theta)$ and $\overline{g}(\theta)$ are bounded by constants $\sigma, \overline{\sigma} > 0$:
    \begin{equation*}
    \begin{array}{cc}
        \E [\Vert g(\theta) - \nabla \gL(\theta) \Vert_2^2] \le \sigma^2, &
         \E[\Vert \overline{g}(\theta) - \nabla \overline{\gL}(\theta) \Vert_2^2] \le \overline{\sigma}^2
    \end{array}
    \end{equation*}
    where $\nabla \gL(\theta)$ and $\nabla \overline{\gL}(\theta)$ are full gradients.
    \label{assumption: variance}
\end{assumption}

The proof is inspired by \citet{wang2019convergence}. Before we prove the theorems, we need the following lemma from~\citep{wang2019convergence} to bound the difference between the gradient of \textit{adversarial loss} and that of \textit{robust loss}.
\begin{lemma}
	Under Assumptions~\ref{assumption: lipschitz} and \ref{assumption: concave}, the approximate stochastic gradient $\hat{g}(\theta)$ satisfies
	\begin{equation}
		\Vert \hat{g}(\theta) - g(\theta) \Vert_2 \le L_{\theta \vx} \sqrt{\frac{\delta}{\mu}}
		\label{eq:lemma1}
	\end{equation}
	\label{lemma:approx_g}
\end{lemma}

\subsubsection{Proof of Theorem~\ref{theorem: same_dir}}
We first prove Theorem~\ref{theorem: same_dir} under the case where $\langle \vm_{1, t} \cdot \vm_{2, t} \rangle > 0$.
\begin{align*}
	\gL(\theta^{t+1}) \le& \gL(\theta^t) + \langle \nabla \gL(\theta^t), \theta^{t+1} - \theta^t \rangle + \frac{L}{2} \Vert \theta^{t+1} - \theta^t \Vert_2^2 \\
	=& \gL(\theta^t) + \langle \nabla \gL(\theta^t), -\eta_t(\frac{1}{\vv_{1, t}} \hat{g}(\theta^t) + \frac{1}{\vv_{2, t}} \overline{g}(\theta^t)) \rangle + \frac{L \eta_t^2}{2} \Vert \frac{1}{\vv_{1, t}} \hat{g}(\theta^t) + \frac{1}{\vv_{2, t}} \overline{g}(\theta^t) \Vert_2^2 \\
	=& \gL(\theta^t) - \frac{\eta_t}{\vv_{1, t}}(1 - \frac{L \eta_t}{2 \vv_{1, t}}) \Vert \gL(\theta^t) \Vert_2^2 + \frac{\eta_t}{\vv_{1, t}}(1 - \frac{L \eta_t}{\vv_{1, t}}) \langle \nabla \gL(\theta^t), \nabla \gL(\theta^t) - \hat{g}(\theta^t) \rangle + \frac{L \eta_t^2}{2 \vv_{1, t}^2} \Vert \hat{g}(\theta^t) - \nabla \gL(\theta^t) \Vert_2^2 \\
	&- \langle \nabla \gL(\theta^t), \frac{\eta_t}{\vv_{2, t}} \overline{g}(\theta_t) \rangle + \frac{L \eta_t^2}{\vv_{1, t} \vv_{2, t}} \hat{g}(\theta^t)^T \overline{g}(\theta^t) + \frac{L \eta_t^2}{2 \vv_{2, t}^2} \Vert \overline{g}(\theta^t) \Vert_2^2 \\
	=& \gL(\theta^t) - \frac{\eta_t}{\vv_{1, t}}(1 - \frac{L \eta_t}{2 \vv_{1, t}}) \Vert \gL(\theta^t) \Vert_2^2 + \frac{\eta_t}{\vv_{1, t}}(1 - \frac{L \eta_t}{\vv_{1, t}}) \langle \nabla \gL(\theta^t), g(\theta^t) - \hat{g}(\theta^t) \rangle \\
	&+ \frac{\eta_t}{\vv_{1, t}}(1 - \frac{L \eta_t}{\vv_{1, t}}) \langle \nabla \gL(\theta^t), \nabla \gL(\theta^t) - g(\theta^t) \rangle + \frac{L \eta_t^2}{2 \vv_{1, t}^2} \Vert \hat{g}(\theta^t) - g(\theta^t) + g(\theta^t) - \nabla \gL(\theta^t) \Vert_2^2 \\
	&- \frac{\eta_t}{\vv_{2, t}} \langle \nabla \gL(\theta^t), \nabla \overline{\gL}(\theta^t) \rangle + \frac{\eta_t}{\vv_{2, t}} \langle \nabla \gL(\theta^t), \overline{\gL}(\theta^t) - \overline{g}(\theta^t) \rangle +
	\frac{L \eta_t^2}{\vv_{1, t} \vv_{2, t}} \hat{g}(\theta^t)^T \overline{g}(\theta^t) + \frac{L \eta_t^2}{2 \vv_{2, t}^2} \Vert \overline{g}(\theta^t) \Vert_2^2 \\
	\le& \gL(\theta^t) - \frac{\eta_t}{2 \vv_{1, t}} \Vert \nabla \gL(\theta^t) \Vert_2^2 + \frac{\eta_t}{2 \vv_{1, t}}(1 + \frac{L \eta_t}{\vv_{1, t}}) \Vert g(\theta^t) - \hat{g}(\theta^t) \Vert_2^2 \\
	&+ \frac{\eta_t}{\vv_{1, t}}(1 - \frac{L \eta_t}{\vv_{1, t}}) \langle \nabla \gL(\theta^t), \nabla \gL(\theta^t) - g(\theta^t) \rangle 
	+ \frac{L \eta_t^2}{\vv_{1, t}^2} (\Vert \hat{g}(\theta^t) - g(\theta^t) \Vert_2^2 + \Vert g(\theta^t) - \nabla \gL(\theta^t) \Vert_2^2) \\
	&- \frac{\eta_t}{\vv_{2, t}} \langle \nabla \gL(\theta^t), \nabla \overline{\gL}(\theta^t) \rangle + \frac{\eta_t}{\vv_{2, t}} \langle \nabla \gL(\theta^t), \overline{\gL}(\theta^t) - \overline{g}(\theta^t) \rangle +
	\frac{L \eta_t^2}{\vv_{1, t} \vv_{2, t}} \hat{g}(\theta^t)^T \overline{g}(\theta^t) + \frac{L \eta_t^2}{2 \vv_{2, t}^2} \Vert \overline{g}(\theta^t) \Vert_2^2	
\end{align*}

Taking expectation on both sides of the above inequality conditioned on $\theta^t$, we have
\begin{align*}
	\E[\gL(\theta^{t+1}) - \gL(\theta^t) | \theta^t] \le&
	-\frac{\eta_t}{2 \Vert \nabla \gL(\theta^t) \Vert_2} \Vert \nabla \gL(\theta^t) \Vert_2^2 + \frac{\eta_t}{2 \Vert \nabla \gL(\theta^t) \Vert_2} (1 + \frac{L \eta_t}{ \Vert \nabla \gL(\theta^t) \Vert_2}) \frac{L_{\theta \vx}^2 \delta}{\mu} \\
	&+ \frac{L \eta_t^2}{\Vert \nabla \gL(\theta^t) \Vert_2^2} \sigma^2 - \frac{\eta_t}{\Vert \nabla \overline{\gL}(\theta^t) \Vert_2} \langle \nabla \gL(\theta^t), \nabla \overline{\gL}(\theta^t) \rangle + \frac{3L \eta_t^2}{2} \\
	=& -\frac{\eta_t}{2} \Vert \nabla \gL(\theta^t) \Vert_2 + \frac{3}{2} L \eta_t^2 + \frac{\eta_t}{2 \Vert \nabla \gL(\theta^t) \Vert_2} (1 + \frac{L \eta_t}{ \Vert \nabla \gL(\theta^t) \Vert_2}) \frac{L_{\theta \vx}^2 \delta}{\mu} \\
	&+ \frac{L \eta_t^2}{\Vert \nabla \gL(\theta^t) \Vert_2^2} \sigma^2 - \frac{\eta_t}{\Vert \nabla \overline{\gL}(\theta^t) \Vert_2} \langle \nabla \gL(\theta^t), \nabla \overline{\gL}(\theta^t) \rangle \\
	=& -\frac{\eta_t}{2} \Vert \nabla \gL(\theta^t) \Vert_2 + \frac{3}{2} L \eta_t^2 + \frac{\eta_t L_{\theta \vx}^2 \delta}{2 \mu \Vert \nabla \gL(\theta^t) \Vert_2} + \frac{L \eta_t^2 L_{\theta \vx}^2 \delta}{2 \mu \Vert \nabla \gL(\theta^t) \Vert_2^2} \\
	&+ \frac{L \eta_t^2}{\Vert \nabla \gL(\theta^t) \Vert_2^2} \sigma^2 - \frac{\eta_t}{\Vert \nabla \overline{\gL}(\theta^t) \Vert_2} \langle \nabla \gL(\theta^t), \nabla \overline{\gL}(\theta^t) \rangle
\end{align*}
Assume the same direction condition $g_\text{adv} \cdot g_\text{IBP} > 0$ holds and $\eta_t = \Vert \nabla \gL(\theta^t) \Vert_2 \cdot \eta$
\begin{equation}
	\E[\gL(\theta^{t+1}) - \gL(\theta^t) | \theta^t] \le
	-\frac{\eta}{2}(1 - 3 L \eta) \Vert \nabla \gL(\theta^t) \Vert_2^2 + \frac{\eta}{2}(1 + L \eta) \frac{L_{\theta \vx} \delta}{\mu} + L \eta^2 \sigma^2
	\label{eq:loss_diff}
\end{equation}
Taking telescope sum of (\ref{eq:loss_diff}) over $t=0,\dots,T-1$, we obtain
\begin{equation*}
	\sum_{t=0}^{T - 1} \frac{\eta}{2} (1 - 3 L \eta) \E[\Vert \nabla \gL(\theta^t) \Vert_2^2] \le \E[\gL(\theta^0) - \gL(\theta^T)] + T \frac{\eta}{2} (1 + L \eta) \frac{L_{\theta \vx}^2 \delta}{\mu} + T L \eta^2 \sigma^2
\end{equation*}

Choosing $\eta = \min(1 / 6 L, \sqrt{\Delta / T L \sigma^2})$, we can show that
\begin{equation*}
	\frac{1}{T} \sum_{t=0}^{T - 1} \E[\Vert \nabla \gL(\theta^t) \Vert_2^2] \le 8 \sigma \sqrt{\frac{L \Delta}{T}} + \frac{7L_{\theta \vx}^2 \delta}{3 \mu}
\end{equation*}

which completes the proof.

\subsubsection{Proof of Theorem~\ref{theorem: diff_dir}}

Now we prove Theorem~\ref{theorem: diff_dir} under the case where $\langle \vm_{1, t} \cdot \vm_{2, t} \rangle \le 0$.

\begin{align*}
	\overline{\gL}(\theta^{t + 1}) \le&
	\overline{\gL}(\theta^t) + \langle \overline{\gL}(\theta^t), \theta^{t+1} - \theta^t \rangle + \frac{\overline{L}}{2} \Vert \theta^{t + 1} - \theta^t \Vert_2^2 \\
	=& \overline{\gL}(\theta^t) + \langle \nabla \overline{\gL}(\theta^t), - \eta_t (1 + \overline{\gL}(\theta^t)) \overline{g}(\theta^t) + \eta_t \frac{\langle \vm_{1, t}, \vm_{2, t} \rangle}{\vv_{1, t}^2} \hat{g}(\theta^t) \rangle \\
	&+ \frac{\overline{L} \eta_t^2}{2} \Vert \frac{\langle \vm_{1, t}, \vm_{2, t} \rangle}{\vv_{1, t}^2} \hat{g}(\theta^t) - (1 + \overline{\gL}(\theta^t)) \overline{g}(\theta^t) \Vert_2^2 \\
	=& \overline{\gL}(\theta^t) - \eta_t (1 + \overline{\gL}(\theta^t)) \langle \nabla \overline{\gL}(\theta^t), \overline{g}(\theta^t) \rangle + \frac{\overline{L} \eta_t^2 (1 + \overline{\gL}(\theta^t))^2}{2} \Vert \overline{g}(\theta^t) \Vert_2^2 \\
	&+ \frac{\eta_t \langle \vm_{1, t}, \vm_{2, t} \rangle}{\vv_{1, t}^2} \langle \nabla \overline{\gL}(\theta^t), \hat{g}(\theta^t) \rangle + \frac{\overline{L} \eta_t^2 \langle \vm_{1, t}, \vm_{2, t} \rangle_2^2}{2 \vv_{1, t}^4} \Vert \hat{g}(\theta^t) \Vert_2^2 \\
	&- \frac{\overline{L} \eta_t^2 \langle \vm_{1, t}, \vm_{2, t} (1 + \overline{\gL}(\theta^t))\rangle}{\vv_{1, t}^2} \langle \hat{g}(\theta^t), \overline{g}(\theta^t) \rangle \\
	=& \overline{\gL}(\theta^t) - \eta_t(1 + \overline{\gL}(\theta^t)) (1 - \frac{\overline{L} \eta_t (1 + \overline{\gL}(\theta^t))}{2}) \Vert \nabla \overline{\gL}(\theta^t) \Vert_2^2 \\
	&+ \eta_t(1 + \overline{\gL}(\theta^t)) (1 - \overline{L} \eta_t (1 + \overline{\gL}(\theta^t))) \langle \nabla \overline{\gL}(\theta^t), \nabla \overline{\gL}(\theta^t) - \overline{g}(\theta^t) \rangle  \\
	&+ \frac{\overline{L} \eta_t^2 (1 + \overline{\gL}(\theta^t))^2}{2} \Vert \nabla \overline{\gL}(\theta^t) - \overline{g}(\theta^t) \Vert_2^2 \\
	&+ \frac{\eta_t \langle \vm_{1, t}, \vm_{2, t} \rangle}{\vv_{1, t}^2} \langle \nabla \overline{\gL}(\theta^t), \hat{g}(\theta^t) \rangle + \frac{\overline{L} \eta_t^2 \langle \vm_{1, t}, \vm_{2, t} \rangle_2^2}{2 \vv_{1, t}^4} \Vert \hat{g}(\theta^t) \Vert_2^2 \\
	&- \frac{\overline{L} \eta_t^2 \langle \vm_{1, t}, \vm_{2, t} (1 + \overline{\gL}(\theta^t))\rangle}{\vv_{1, t}^2} \langle \hat{g}(\theta^t), \overline{g}(\theta^t) \rangle
\end{align*}

Taking expectation on both sides of the above inequality conditioned on $\theta^t$, we have
\begin{equation}
	\begin{aligned}
		\E[\overline{\gL}(\theta^{t+1}) - \overline{\gL}(\theta^t) | \theta^t] \le&
		- \eta_t(1 + \overline{\gL}(\theta^t)) (1 - \frac{\overline{L} \eta_t (1 + \overline{\gL}(\theta^t))}{2}) \Vert \nabla \overline{\gL}(\theta^t) \Vert_2^2 + \frac{\overline{L} \eta_t^2 (1 + \overline{\gL}(\theta^t))^2}{2} \overline{\sigma}^2 \\
		&+ \eta_t \Vert \nabla \overline{\gL}(\theta^t) \Vert_2^2 + \frac{1}{2} \overline{L} \eta_t^2 \Vert \nabla \overline{\gL}(\theta^t) \Vert_2^2 \\
		=& - \eta_t  (\overline{\gL}(\theta^t) - \frac{\overline{L} \eta_t}{2}) \Vert \nabla \overline{\gL}(\theta^t) \Vert_2^2 + \frac{\overline{L} \eta_t^2 (1 + \overline{\gL}(\theta^t))^2}{2} \overline{\sigma}^2
	\end{aligned}
	\label{eq:opposite_loss_diff}
\end{equation}

Taking telescope sum of~\ref{eq:opposite_loss_diff} over $t=0, \dots, T - 1$, we obtain
\begin{equation*}
	\sum_{t = 0}^{T - 1} \eta_t (\E[\overline{\gL}(\theta^t)] - \frac{\overline{L} \eta_t}{2}) \E[\Vert \nabla \overline{\gL}(\theta^t) \Vert_2^2] \le \E[\overline{\gL}(\theta^0) - \overline{\gL}(\theta^T)] + \sum_{t = 0}^{T - 1} \frac{\overline{L}\eta_t^2 (1 + \E[\overline{\gL}(\theta^t)])^2}{2} \overline{\sigma}^2
\end{equation*}

Choosing $\eta_t = \min( 2 * \E[\overline{\gL}(\theta^t)] / \overline{L} - 1 / \overline{L}, \sqrt{\Delta / T \overline{L} \sigma^2})$, if $\E[\overline{\gL}(\theta^t)] > 1 / 2$, then we can show that

\begin{equation*}
	\frac{1}{T} \sum_{t=0}^{T - 1} \E[\Vert \nabla \overline{\gL}(\theta^t) \Vert_2^2] \le 2
\end{equation*}

\end{strip}
\noindent
which completes the proof.